\documentclass{article}

\PassOptionsToPackage{round,authoryear}{natbib}
\usepackage[main,final]{neurips_2026}

\usepackage[utf8]{inputenc} 
\usepackage[T1]{fontenc}    
\usepackage{hyperref}       
\usepackage{url}            
\usepackage{booktabs}       
\usepackage{amsfonts}       
\usepackage{microtype}      
\usepackage[table]{xcolor} 

\usepackage{graphicx}
\usepackage{caption}
\usepackage{tcolorbox}

\usepackage{algorithm}
\usepackage{algorithmic}
\usepackage{amsmath}
\usepackage{amssymb}

\usepackage[capitalize,noabbrev]{cleveref}

\usepackage{multirow}
\usepackage{xurl} 

\title{Counterfactual Distillation: Internalizing Reflective Experience into LLM Agent}

\author{%
  \normalfont
  \makebox[0pt][c]{%
  \parbox[t]{\textwidth}{%
    \raggedright
    \mbox{\textbf{Rui Ge}\textsuperscript{1}\thanks{Work done during an internship at the University of California San Diego.}}\quad
    \mbox{\textbf{Yichao Fu}\textsuperscript{1}}\quad
    \mbox{\textbf{Yu-Yang Qian}\textsuperscript{2}}\quad
    \mbox{\textbf{Junda Su}\textsuperscript{1}}\quad
    \mbox{\textbf{Yiming Zhao}\textsuperscript{1}}\quad
    \mbox{\textbf{Hao Zhang}\textsuperscript{1}\thanks{Corresponding Author: Hao Zhang \textless\href{mailto:haozhang@ucsd.edu}{haozhang@ucsd.edu}\textgreater.}}\\
    \textsuperscript{1}University of California San Diego\quad
    \textsuperscript{2}Nanjing University
  }%
  }
}

\begin{document}

\maketitle

\begin{abstract}
Large language models are increasingly deployed as autonomous agents that plan and act over long-horizon interactions, learning from both final outcomes and rich environmental feedback. However, optimizing from such interaction experience remains challenging, as feedback is heterogeneous and rarely carries explicit reward information. Existing post-training approaches rely on outcome-driven methods, such as reinforcement learning with verifiable rewards (RLVR), which primarily exploit final success signals while leaving interaction experience and feedback underutilized. In sparse-reward, long-horizon tasks, this often results in \emph{distribution sharpening}: the policy reinforces a narrow set of already-successful behaviors, without substantially improving the feedback-grounded agency needed for broader problem-solving capacity(e.g., Pass@$k$). 
We propose \emph{Counterfactual Distillation}, a framework that converts exploration-derived experience into trainable policy supervision. Specifically, we organize exploration as a tree-structured search process, where the agent reflects on past decision points and makes new experience-guided decisions to form alternative branches. These corrections are then distilled as counterfactual targets: we train the model to predict the revised action from the original history without explicitly providing the experience, thereby introducing out-of-distribution behavior patterns and expanding exploration capacity. Across diverse interactive coding and agentic tasks, our method outperforms outcome-driven baselines such as GRPO and experience-based methods such as Early Experience, with gains of up to 14 percentage points over the base model on Pass@128. When interleaved with reinforcement learning updates, it further raises the performance ceiling, improving Pass@1 on Sokoban by 8 percentage points over extended GRPO.
Our code is available at \url{https://github.com/motigrez/Counterfactual-Distillation}.
\end{abstract}

\section{Introduction}

Large Language Models (LLMs) are increasingly deployed as autonomous agents that plan, act, and adapt over \emph{long-horizon, multi-turn} interactions with complex environments~\citep{yao2022react,nakano2021webgpt}. In such settings, success depends less on raw single-shot reasoning and more on the agent's ability to understand and adapt to the environment, since many failures arise from unfamiliarity with the rules or recurring pitfalls of the environment~\citep{yang2024swe,shinn2023reflexion}. This makes \emph{learning from experience} a central capability for long-horizon agents, allowing them to convert lessons from past failures into actionable knowledge for future attempts. A growing body of work exploits this idea at test time: agents accumulate insights into evolving playbooks or external knowledge bases for in-context reuse~\citep{zhang2025agentic}, or perform inference-time rollback and search to repair failing trajectories~\citep{li2025generator}. While effective, these approaches rely on substantial test-time compute or external memory infrastructure, which is often impractical under latency- or cost-constrained deployment, where each query may afford only a single rollout.

\begin{figure}[t]
  \centering
  \includegraphics[width=0.5\linewidth]{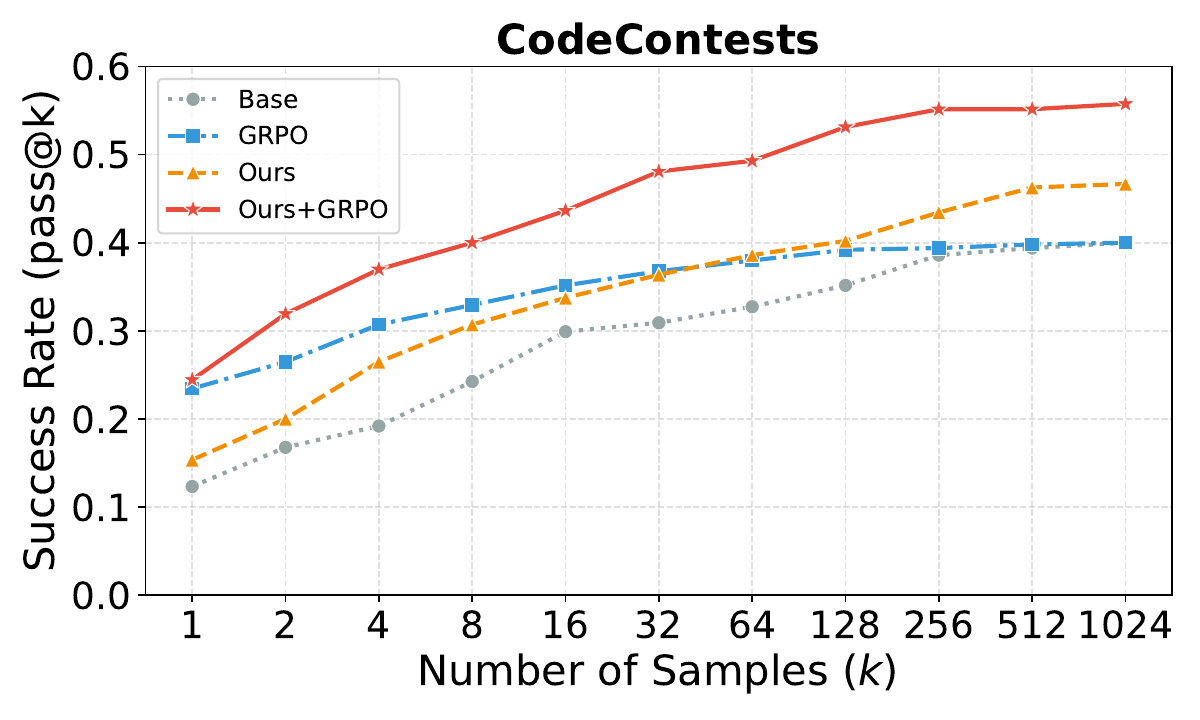} 
  \caption{Internalizing feedback-grounded agency improves model capability (i.e., Pass@K) in long-horizon interaction, while outcome-only training (e.g., GRPO) yields limited gains beyond the base model.}
  \label{fig:teaser_capability}
\end{figure}

A widely adopted alternative is reinforcement learning with verifiable rewards (RLVR)~\citep{guo2025deepseek,zeng2025simplerl,shao2024deepseekmath}, which directly optimizes the policy from a terminal outcome reward and reduces reliance on test-time scaling. Despite its appealing simplicity, RLVR has several notable limitations in long-horizon agentic settings. First, it collapses the rich, structured environmental feedback (e.g., compiler errors, invalid actions, intermediate observations) into a single scalar at the episode level, leaving most interaction information unused. Second, rollouts are treated as independent samples scored only by their terminal reward, with no mechanism for accumulating or transferring experience across attempts. Third, sparse rewards make credit assignment difficult in long-horizon tasks: without reliable step-level signals, exploratory deviations cannot be properly attributed or reinforced, and optimization is biased toward behaviors already supported by the model. As a consequence, RLVR tends to behave like \emph{distribution sharpening}~\citep{yue2025does,zhao2025echo}: it concentrates probability mass on already-successful behaviors latent in the base model, boosting Pass@1 while yielding limited or even negative gains at large $k$ (e.g., Pass@1024). This limitation is most visible under Pass@k evaluation: Pass@1 reflects single-rollout effectiveness, whereas Pass@k at large $k$ approximates the model's true capability ceiling by exposing whether successful strategies exist anywhere in its distribution. As previewed in Figure~\ref{fig:teaser_capability}, our method substantially expands this ceiling in long-horizon interaction, while RLVR offers little improvement over the base model. 

These limitations point to an underused source of supervision in agentic training. Environmental feedback accumulated over long-horizon interaction contains far richer information than a terminal scalar reward. The key challenge is how to internalize this experience during training, as such experience is often heterogeneous, free-form, and structurally misaligned with the policy's action space: agentic tasks ultimately require grounded \emph{think-then-act} interactions with the environment, rather than natural language commentary. Consequently, naively conditioning on or imitating natural-language experience does not by itself enrich the policy's intrinsic action distribution. Our goal is therefore to convert exploration-derived experience into \emph{trainable policy supervision} that can be internalized into model parameters, thereby expanding the policy's exploration space and raising its capability ceiling without requiring explicit experience or heavy test-time scaling at inference.

To this end, we propose \textbf{Counterfactual Distillation}, a framework that converts reflective experience into trainable policy supervision through a two-stage pipeline. In Stage~1 (\S\ref{subsec:tree_search}, \emph{Tree-Structured Exploration for Experience Generation}), the agent reflects on its trajectory, localizes a suboptimal step $\tau$, and synthesizes a brief natural-language experience $e$ that diagnoses the failure and proposes a fix; after restoring the environment back to $\tau$, it re-decides under the experience-augmented context, producing a revised action $a'_\tau$. By consuming the structured feedback in place and propagating diagnoses across attempts through a search tree, this stage turns the rich interaction signal into experience that accumulates across rollouts. In Stage~2 (\S\ref{subsec:cf_distillation}, \emph{Experience Distillation}), we strip $e$ from the context and train the model to reproduce $a'_\tau$ from the original history $h_\tau$ alone. The objective is \emph{counterfactual} in the sense that it uses the experience-guided revision as supervision for the original decision context, training the model to behave as though the information in $e$ had already been internalized. Each training pair therefore transfers the behavioral change induced by $e$ back to the original decision context, increasing the probability of feedback-grounded alternatives that were less likely under the original policy. In this way, Counterfactual Distillation converts an explicit in-context intervention into an implicit shift of the policy distribution, yielding localized decision-level supervision and expanding the behavioral modes accessible at inference.

We evaluate our method across a diverse suite of long-horizon agentic benchmarks, including CodeContests~\citep{codecontest}, WebShop~\citep{yao2022webshop}, ALFWorld~\citep{shridhar2020alfworld}, ScienceWorld~\citep{wang2022scienceworld}, and Sokoban~\citep{SchraderSokoban2018}. Across tasks and model scales, our method consistently outperforms outcome-driven RLVR baselines such as GRPO~\citep{guo2025deepseek} as well as experience-based methods such as Early Experience\citep{zhang2025agent}, with gains of up to 13.94 percentage points over the base model on Pass@128(\Cref{tab:main_results,tab:codecontest_results}). Moreover, in the on-policy interleaving experiment (Table~\ref{tab:grpo_training_ceiling}), Counterfactual Distillation further raises the \emph{optimization ceiling} of outcome-driven RL: after additional GRPO updates have largely saturated, interleaving our distillation cycles continues to improve the performance, improving Pass@1 on Sokoban by 8.00 percentage points over extended GRPO. Together, these results indicate that counterfactually distilling exploration-derived experience provides a complementary axis to scalar-reward optimization: it injects structured, behaviorally diverse supervision that pure RLVR cannot easily access.

To summarize, our contributions can be listed as follows:
\begin{itemize}
    \item We propose \textbf{Counterfactual Distillation}, a framework for internalizing reflective experience into the policy, enabling agents to benefit from experience without heavy test-time scaling at inference.

    \item We provide a complementary learning signal to outcome-driven RLVR, mitigating its reliance on terminal rewards by reusing rich interaction feedback, transferring experience across attempts, and supplying localized decision-level supervision.

    \item We conduct extensive experiments across diverse long-horizon agentic benchmarks, showing consistent improvements over RLVR and experience-based baselines, especially in large-$k$ performance, and further gains when interleaved with on-policy GRPO training.
\end{itemize}

\section{Background}

\subsection{LLMs as Interacting Agents}

Following ReAct~\citep{yao2022react}, we model an interactive episode as an interleaved sequence of observations and actions. Given an instruction $q\in\mathcal{Q}$, at step $t$ the environment returns a textual observation $o_t$, and the LLM policy $\pi_\theta$ generates an executable action
\begin{equation}
\label{eq:action}
a_t \sim \pi_\theta(\cdot \mid h_t, q), \qquad
h_t=(o_0,a_0,\ldots,o_{t-1},a_{t-1},o_t).
\end{equation}
A rollout is denoted by $\tau=(q,o_0,a_0,\ldots,o_T)$, whose outcome is evaluated by a task-specific verifier or environment as a success/failure signal.

\subsection{Reinforcement Learning with Verifiable Rewards}
A common post-training paradigm for agentic LLMs is RL with verifiable rewards (RLVR), where a verifier or environment assigns a scalar reward to a completed trajectory, e.g., success/failure or the number of passing tests. Representative algorithms include PPO~\citep{schulman2017proximal} and group-based methods such as GRPO~\citep{guo2025deepseek,zeng2025simplerl,shao2024deepseekmath}.

Given $G$ sampled traces for the same problem, GRPO computes normalized advantages within the group and optimizes
\begin{align*}
\ell_i(\theta)
&=
\min\!\Big(
r_i(\theta)A_i,
\mathrm{clip}(r_i(\theta),1-\epsilon,1+\epsilon)A_i
\Big)
-\beta\,\mathrm{KL}, \\
J_{\mathrm{GRPO}}(\theta)
&=
\mathbb{E}_{q,\{o_i\}}\!\left[
\frac{1}{G}\sum_{i=1}^{G}\ell_i(\theta)
\right],
\end{align*}
where $\mathrm{KL}\triangleq D_{\mathrm{KL}}(\pi_\theta\|\pi_{\mathrm{ref}})$.
Although effective, outcome-based RLVR provides only trajectory-level supervision, making credit assignment difficult in long-horizon generation and tool use. Since updates are driven by rewards on sampled rollouts, they often reweight or sharpen behaviors within the model's existing solution space, rather than reliably discovering qualitatively new solutions. This motivates evaluating capability under larger sampling budgets.

\subsection{LLM's Capability Boundary}\label{sec:metrics}
To quantify a model's capability boundary, we evaluate Pass@$K$ following prior work~\citep{yue2025does}. For each problem, we sample $K$ rollouts and assign score $1$ if any rollout succeeds under the task-specific verifier/environment, and $0$ otherwise. Pass@$K$ is the average score over the dataset.

Pass@$1$ measures single-rollout effectiveness, which is most relevant for deployment, while large-$K$ Pass@$K$ estimates the model's best-of-$K$ potential, i.e., whether successful strategies exist within its rollout distribution. In our experiments, we use Pass@$128$ to characterize the empirical capability boundary and additionally report Pass@$1$ for single-attempt performance.

\section{Counterfactual Distillation: Learning From Reflective Experience}
\label{sec:cd}
 
\begin{figure*}
    \centering
    \includegraphics[width=0.98\linewidth]{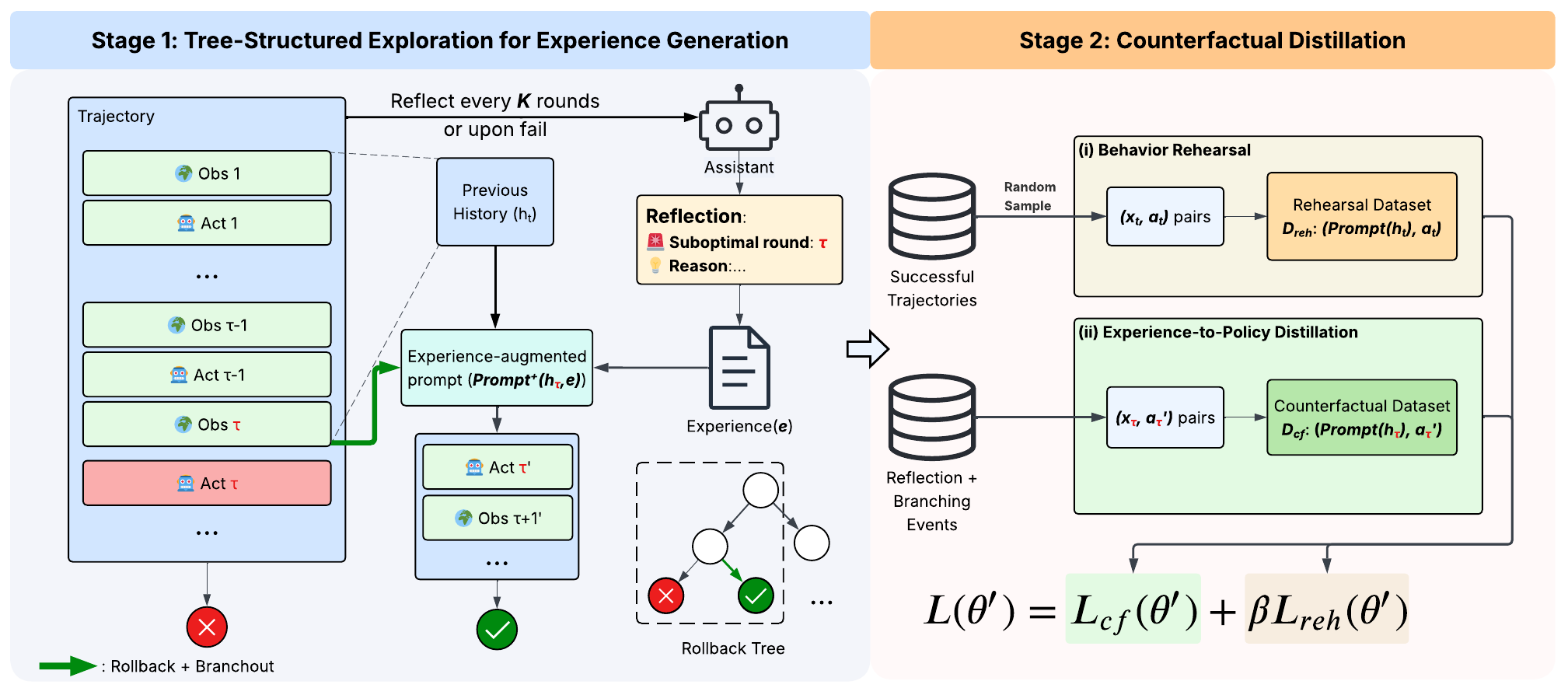}
    \caption{Illustration of the Counterfactual Distillation framework. \textbf{Stage 1}: During experience collection, the assistant periodically reviews the current trajectory and identifies a suboptimal round (denoted as red-colored $\tau$). It then produces the actionable experience $e$, which is concatenated with the restored history to facilitate subsequent attempts. \textbf{Stage 2}: During experience distillation, the model optimizes a joint loss using two datasets: randomly sampled rehearsal pairs to maintain capabilities, and counterfactual pairs (original prompts paired with experience-improved actions) to internalize diverse exploration. For simplicity, we depict one branching event from the rollback exploration tree.}
    \label{fig:lefe-overview}
\end{figure*}
This section introduces Counterfactual Distillation, a two-stage framework that converts rollout-time experience into counterfactual supervision for policy learning (see Figure~\ref{fig:lefe-overview}). 
Stage~1 is the rollout stage (\S\ref{subsec:tree_search}), where the agent interacts with the environment, reflects on trajectories, and performs tree-structured search to collect reflective experience and experience-guided revision actions. The pseudo-code is shown in Algorithm~\ref{alg:lefe_generation}. 
Stage~2 is the training stage (\S\ref{subsec:cf_distillation}), where we remove the explicit experience and train the model to reproduce the revised actions from the original decision contexts, thereby internalizing the experience-induced policy shift into model parameters.

\subsection{Tree-Structured Exploration for Experience Generation}
\label{subsec:tree_search}
 
\paragraph{Setup and Notation.}
We consider an episodic interactive environment $\mathcal{E}$ with horizon $T$.
For an instruction $q \in \mathcal{Q}$, an episode is a sequence of states $\{E_t\}_{t=0}^{T}$; at each step, the agent observes $o_t$ and emits an executable action $a_t$. A language-model policy $\pi_\theta$ induces a distribution over actions conditioned on history as in Eq.~\ref{eq:action}, with environment dynamics $E_{t+1}, o_{t+1} \leftarrow \mathrm{Step}(E_t, a_t)$.
A \emph{rollout} executes the policy from an initial state until termination (success, irreversible failure, or step limit).

\paragraph{Reflection and Experience Generation.}
A fundamental challenge in leveraging experience is that linguistic feedback is qualitative and unstructured, and thus difficult to internalize directly through standard optimization.
We address this by treating experience as a \emph{contextual intervention} that induces a \emph{policy shift} via the input context.
Concretely, every $K$ steps or upon failure, the agent invokes a reflection procedure. Given the interaction history $h_t$ and a reflection prompt $p_{\text{refl}}$, the policy $\pi_\theta$ produces a target step $\tau$ and a natural-language experience $e$:
\begin{equation*}
(\tau, e) \sim \pi_\theta(\cdot \mid h_t, p_{\text{refl}}), \quad \tau \in \{1, \dots, t\},
\end{equation*}
where $\tau$ localizes the step at which the trajectory deviated from the desired path, and $e$ provides a diagnostic with actionable suggestions for re-deciding at $\tau$.
 
\paragraph{Branching via Replay.}
To explore an alternative branch from $(\tau, e)$, we restore the environment state $E_\tau$ and the corresponding history $h_\tau$ by resetting $\mathcal{E}$ and replaying the original action sequence $a_{1:\tau-1}$.
Conditioned on the synthesized experience $e$, the policy then produces a revised action and continues the rollout:
\begin{equation*}
a_\tau' \sim \pi_\theta(\cdot \mid h_\tau, q, e), \qquad E_{\tau+1}', o_{\tau+1}' \leftarrow \mathrm{Step}(E_\tau, a_\tau').
\end{equation*}
The agent proceeds from $E_{\tau+1}'$ until the next reflection or termination. We organize branching requests in a queue and expand them in a breadth-first manner, yielding an implicit \emph{experience tree} whose nodes are decision points and whose edges are experience-guided revisions. Expansion stops when reaching a maximum depth or exhausting the attempt budget.
 
Figure~\ref{fig:lefe-example} provides a concrete example from Sokoban illustrating reflection, state restoration via replay, and branch exploration in practice.

\subsection{Counterfactual Distillation}
\label{subsec:cf_distillation}
 
Stage~1 yields, for each branching event, a triple $(h_\tau, e, a_\tau')$: an original history $h_\tau$, a natural-language experience $e$, and an experience-guided action $a_\tau'$ that is \emph{counterfactual} to the action originally taken at step $\tau$. Stage~2 internalizes these counterfactual revisions into the policy parameters.
 
\paragraph{(i) Counterfactual Objective.}
The core supervision is derived from the branching events identified during reflection.
Whereas $a_\tau'$ was generated under the experience-augmented context $(h_\tau, q, e)$, we ask the policy to reproduce $a_\tau'$ from the experience-\emph{free} context $(h_\tau, q)$ alone:
\begin{equation*}
\mathcal{L}_{\text{cf}}(\theta') = -\mathbb{E}_{(h_\tau, a_\tau') \sim \mathcal{D}_{\text{cf}}} \left[ \log \pi_{\theta'}(a_\tau' \mid h_\tau, q) \right].
\end{equation*}
This objective is counterfactual in two complementary senses. At the \emph{decision level}, it commits the policy to $a_\tau'$ at $h_\tau$ as though the lesson encoded in $e$ were already part of its own knowledge, removing the need for explicit reflection at inference. At the \emph{distribution level}, the in-context experience acts as a lever that lifts a previously low-probability action; training on $(h_\tau, a_\tau')$ alone absorbs this lift into the policy parameters as a persistent shift in support. As illustrated in Figure~\ref{fig:lefe-after-train-example}, when the trained model later encounters the same situation, it avoids repeating the original mistake without receiving any explicit experience.
 
\paragraph{(ii) Behavior Rehearsal.}
To preserve the agent's foundational competence and mitigate catastrophic forgetting under the counterfactual update, we additionally rehearse on successful trajectories. Following rejection-sampling-style imitation~\citep{ahn2024large}, we treat successful rollouts---including those discovered through branching---as high-quality demonstrations. For each successful trajectory, we extract state-action pairs $(h_t, a_t)$ and minimize:
\begin{equation*}
\mathcal{L}_{\text{reh}}(\theta') = -\mathbb{E}_{(h, a) \sim \mathcal{D}_{\text{reh}}} \left[ \log \pi_{\theta'}(a \mid h, q) \right].
\end{equation*}
 
\paragraph{Training Objective.}
The two losses are jointly optimized:
\begin{equation}
\mathcal{L}(\theta') = \mathcal{L}_{\text{cf}}(\theta') + \beta\, \mathcal{L}_{\text{reh}}(\theta'),
\label{eq:cd_loss}
\end{equation}
where $\beta$ balances rehearsal strength. $\mathcal{L}_{\text{cf}}$ injects experience-induced alternatives into the policy's intrinsic action distribution---expanding behavioral coverage and improving large-$k$ metrics such as pass@$k$---while $\mathcal{L}_{\text{reh}}$ retains the model's existing task competence.

\subsection{On-Policy Counterfactual Distillation.}
\label{on-policy}
Our method naturally splits into a \emph{rollout} phase (Stage~1) that collects experience under the current policy and an \emph{update} phase (Stage~2) that trains on it, mirroring the structure of standard RL. We exploit this parallel and propose an on-policy variant of Counterfactual Distillation that composes cleanly with policy-gradient methods.

In each on-policy cycle, the rollout phase uses the latest policy $\pi_\theta$ to grow a fresh experience tree, recording both branching events $(h_\tau, a_\tau')$ and the reflection acts $(h_t, \tau, e)$ that produced them. The update phase then reinforces two kinds of supervision: (a) the revised action $a_\tau'$ via $\mathcal{L}_{\text{cf}}$, as before, and (b) reflection acts whose induced branches eventually succeed---i.e., reflections that proved productive in hindsight. Letting $\mathcal{D}_{\text{refl}}^{+}$ denote such successful reflections, we add:
\begin{equation*}
\mathcal{L}_{\text{refl}}(\theta') = -\mathbb{E}_{(h_t, \tau, e) \sim \mathcal{D}_{\text{refl}}^{+}}\!\left[\log \pi_{\theta'}(\tau, e \mid h_t, p_{\text{refl}})\right],
\end{equation*}
so that the policy learns not only \emph{how to revise} but also \emph{when and how to reflect}. The full on-policy objective is $\mathcal{L}_{\text{cf}} + \beta\,\mathcal{L}_{\text{reh}} + \gamma\,\mathcal{L}_{\text{refl}}$.

On-policy Counterfactual Distillation can be directly interleaved with standard RL: every $M$ GRPO steps, we insert one cycle and then resume RL from the updated checkpoint. As shown in Table~\ref{tab:grpo_training_ceiling}, under the same extended training budget where standalone GRPO has plateaued, this interleaved schedule continues to deliver gains, breaking through the ceiling of outcome-driven RL.

\section{Experiment}
\label{sec:experiments}
We evaluate Counterfactual Distillation (CD) across a diverse range of agentic tasks, spanning competitive programming and multi-step interactive reasoning in simulated physical and digital environments.

\textbf{Models.} We evaluate Counterfactual Distillation using the Qwen2.5 (7B/72B)~\citep{team2024qwen2} and Llama-3/3.1 (8B/70B)~\citep{grattafiori2024llama} series. These models are widely adopted as standardized backbones in recent RLVR~\citep{zeng2025simplerl,shao2025spurious} and agentic reasoning~\citep{da2025agent,feng2025group} research.

\textbf{Datasets.} We evaluate on four interactive benchmarks that demand strategic interaction, long-horizon decision making, and mistake recovery:
(1) \emph{WebShop}~\citep{yao2022webshop}, which requires multi-hop search and attribute matching across over one million products;
(2) \emph{ALFWorld}~\citep{shridhar2020alfworld}, a text-based interactive environment aligned with embodied household tasks;
(3) \emph{ScienceWorld}~\citep{wang2022scienceworld}, a text-based scientific experimentation environment requiring multi-step manipulation under procedural constraints; and
(4) \emph{Sokoban}~\citep{SchraderSokoban2018}, a planning-intensive puzzle environment requiring precise sequential manipulation and correction of earlier mistakes.
We additionally evaluate on \emph{CodeContests}~\citep{codecontest}, a competitive programming benchmark with execution-based test-case verification. Together, these benchmarks cover web navigation, embodied interaction, scientific experimentation, and program synthesis.

\textbf{Baselines \& Metrics.}
We compare Counterfactual Distillation against four baselines: \emph{Base} (instruct model with no task-specific fine-tuning), \emph{GRPO-RLVR} (outcome-supervised RL with verifiable final rewards via GRPO~\citep{guo2025deepseek}), \emph{EarlyExp} (reward-free agent learning that converts early interaction experience into supervision~\citep{zhang2025agent}), and \emph{ACE} (a training-free, prompt-based method that improves agents via evolving playbooks from execution feedback~\citep{zhang2025agentic}).

We report Pass@1 and Pass@128 (Sec.~\ref{sec:metrics}). Pass@1 quantifies single-try pass rate as a measure of exploitation, while Pass@128 estimates best-of-$k$ performance under a much larger sampling budget, serving as a proxy for exploration capacity and the model's fundamental performance boundary. All reported Pass@128 values are computed from \emph{independent} inference-time samples from the trained policy (i.e., without Stage~1 rollback or experience-guided branching). For Qwen2.5-7B-Instruct, Appendix~\ref{app:eval_variability} reports mean $\pm$ SD from 32 complete Pass@$1$ evaluations and 32 Pass@$128$ subsamples of fixed rollout pools.

\begin{table*}[t]
\centering
\caption{Main results on four agent benchmarks. We report Pass@1(\%) and Pass@128(\%) for two models per dataset. Counterfactual Distillation consistently outperforms baselines across all tasks in terms of Pass@128.}
\vspace{-1mm}
\label{tab:main_results}
\begin{small}
\begin{sc}
\setlength{\tabcolsep}{4pt}
\resizebox{\textwidth}{!}{%
\begin{tabular}{l cccc c cccc}
\toprule
& \multicolumn{4}{c}{\textbf{WebShop}} & & \multicolumn{4}{c}{\textbf{AlfWorld}} \\
\cmidrule(lr){2-5} \cmidrule(lr){7-10}
& \multicolumn{2}{c}{Qwen2.5-7B} & \multicolumn{2}{c}{Llama3.1-8B} & & \multicolumn{2}{c}{Qwen2.5-7B} & \multicolumn{2}{c}{Llama3.1-8B} \\
\cmidrule(lr){2-3} \cmidrule(lr){4-5} \cmidrule(lr){7-8} \cmidrule(lr){9-10}
Method & Pass@1 & Pass@128 & Pass@1 & Pass@128 & & Pass@1 & Pass@128 & Pass@1 & Pass@128 \\
\midrule
Base (No FT) & 0.05 & 5.20 & 0.00 & 1.80 & & 26.07 & 78.57 & 29.82 & 74.29 \\
EarlyExp     & 61.55 & 84.60 & 51.13 & 77.80 & & \textbf{72.23} & 92.86 & \textbf{74.15} & 95.71 \\
GRPO-RLVR    & 67.45 & 85.40 & 54.95 & 79.40 & & 69.46 & 91.43 & 72.50 & 94.29 \\
ACE          & \textbf{68.65} & 86.80 & 54.35 & 79.80 & & 66.34 & 89.63 & 70.32 & 92.50 \\
\rowcolor[gray]{0.95}
Ours        & 66.50 & \textbf{89.20} & \textbf{56.25} & \textbf{83.20} & & 67.50 & \textbf{94.29} & 71.79 & \textbf{96.43} \\
\midrule
\midrule
& \multicolumn{4}{c}{\textbf{SciWorld}} & & \multicolumn{4}{c}{\textbf{Sokoban}} \\
\cmidrule(lr){2-5} \cmidrule(lr){7-10}
& \multicolumn{2}{c}{Qwen2.5-7B} & \multicolumn{2}{c}{Llama3.1-8B} & & \multicolumn{2}{c}{Qwen2.5-7B} & \multicolumn{2}{c}{Llama3.1-8B} \\
\cmidrule(lr){2-3} \cmidrule(lr){4-5} \cmidrule(lr){7-8} \cmidrule(lr){9-10}
Method & Pass@1 & Pass@128 & Pass@1 & Pass@128 & & Pass@1 & Pass@128 & Pass@1 & Pass@128 \\
\midrule
Base (No FT) & 7.00 & 47.33 & 7.17 & 48.67 & & 6.90 & 43.80 & 17.70 & 61.40 \\
EarlyExp     & 26.17 & 54.67 & 24.04 & 56.00 & & 60.15 & 71.60 & 57.32 & 68.20 \\
GRPO-RLVR    & 27.17 & 57.33 & 24.25 & 56.00 & & 58.15 & 68.00 & 60.43 & 73.40 \\
ACE          & \textbf{28.45} & 50.00 & \textbf{25.28} & 57.33 & & 61.30 & 70.80 & 60.79 & 73.20 \\
\rowcolor[gray]{0.95}
Ours        & 27.88 & \textbf{62.00} & 22.70 & \textbf{59.33} & & \textbf{64.60} & \textbf{78.40} & \textbf{62.00} & \textbf{77.20} \\
\bottomrule
\end{tabular}%
}
\end{sc}
\end{small}
\vskip -0.1in
\vspace{1mm}
\end{table*}

\textbf{Implementation Details.}
For the interactive agentic benchmarks, we use \emph{verl-agent}~\citep{feng2025group} as a unified framework for both training and evaluation. We adopt the framework-provided settings and interfaces for \emph{ALFWorld}, \emph{WebShop}, and \emph{Sokoban}, and implement a compatible verl-agent wrapper for \emph{ScienceWorld}. For \emph{CodeContests}, we train with \emph{verl}~\citep{sheng2025hybridflow} and follow \emph{CTRL}~\citep{xie2025teaching} for execution-based evaluation.

For baselines, \emph{Base} is the instruction-tuned model without task-specific training, and \emph{GRPO-RLVR} applies task-specific GRPO with verifiable rewards. For \emph{EarlyExp}, since no official codebase is available, we re-implement its Implicit World Modeling (IWM) component via the \emph{verl} SFT module followed by GRPO training. We use the IWM variant for the main comparison because EarlyExp-SR relies on expert demonstrations, whereas our setting learns from model-generated interaction and environment feedback; results with EarlyExp-SR are reported separately in Table~\ref{tab:earlyexp_sr_comparison}. For \emph{ACE}, we extend its official \emph{AppWorld} playbook-based prompting strategy to our benchmarks and apply it on top of the GRPO-trained model under the same \emph{verl-agent} protocol. Our method is initialized from the GRPO-trained model and distills reflective experience data. More details are in Appendix~\ref{implementation-detals}.

\subsection{Main Results}\label{sec:main-result}

\paragraph{Large-$k$ capability gains.}
Table~\ref{tab:main_results} summarizes results on WebShop, ALFWorld, ScienceWorld, and Sokoban across Qwen2.5 and Llama3.1 models. Table~\ref{tab:codecontest_results} further reports results on CodeContests with larger backbones. We report both Pass@1 and Pass@128 to distinguish single-rollout performance from broader capability under increased sampling.

Our advantage is most pronounced at large $k$. While GRPO sometimes matches or exceeds Counterfactual Distillation on Pass@1, its gains plateau as the sampling budget grows, whereas our method consistently achieves stronger Pass@128, indicating broader behavioral coverage. For example, on WebShop with Qwen2.5-7B, GRPO obtains a higher Pass@1, but Counterfactual Distillation achieves a higher Pass@128. This is consistent with \emph{distribution sharpening}: GRPO increases the likelihood of already-successful trajectories, but provides limited additional coverage as $k$ grows. The contrast is especially clear on CodeContests, where Counterfactual Distillation improves Pass@128 by up to \textbf{14 points} over the base model (11 points over GRPO) with Qwen2.5-72B. Overall, these results show that our method better internalizes feedback-grounded agency and expands the model's effective capability in long-horizon interaction.

\paragraph{Raising the training ceiling of GRPO.}
To study the ceiling of GRPO in long-horizon interactive tasks, we evaluate on-policy Counterfactual Distillation(\S\ref{on-policy}) interleaved with RL (RL+CD) on WebShop and Sokoban. Starting from the GRPO-RLVR checkpoint in the main table, we continue training Qwen2.5-7B-Instruct with GRPO for 100 extra steps on WebShop (200 total) and 150 extra steps on Sokoban (300 total). In RL+CD, every 50 GRPO steps we insert 5 on-policy Counterfactual Distillation cycles (data collection and training), then resume GRPO from the updated checkpoint.

As shown in Table~\ref{tab:grpo_training_ceiling}, extended GRPO training improves the success rate to 71.22\% on WebShop and 70.70\% on Sokoban, close to the observed ceiling, as further GRPO updates bring little additional gain. In comparison, RL+CD raises performance to 74.33\% and 78.70\%, respectively. For Pass@128, extended GRPO reaches 88.60\% on WebShop and 75.20\% on Sokoban, while RL+CD achieves 91.00\% and 81.20\%, respectively. These results suggest that late-stage GRPO is limited by sparse rewards and existing decision patterns, making it difficult to induce broader exploration through more RL steps alone. By distilling exploration-derived experience into reflection and revised decisions, on-policy Counterfactual Distillation injects feedback-grounded corrective strategies into the training loop, expanding the effective exploration space and raising the attainable ceiling of outcome-driven RL.

\begin{table}[t]
\centering
\caption{Main results on CodeContests with larger backbone models. Ours uses GRPO$\rightarrow$CD. Counterfactual Distillation yields substantial improvements in Pass@128 over the GRPO-RLVR baseline. \textit{Note:} EarlyExp and ACE are not reported on CodeContests because it is designed for interactive environments and does not directly apply to code execution feedback without a different implementation.}
\label{tab:codecontest_results}
\vspace{2mm}
\begin{small}
\begin{sc}
\setlength{\tabcolsep}{4pt}
\renewcommand{\arraystretch}{0.95}
\begin{tabular}{@{}lcccc@{}}
\toprule
& \multicolumn{2}{c}{Qwen2.5-72B}
& \multicolumn{2}{c}{Llama3-70B} \\
\cmidrule(lr){2-3} \cmidrule(lr){4-5}
Method & Pass@1 & Pass@128 & Pass@1 & Pass@128 \\
\midrule
Base (No FT) & 10.00 & 33.94 & 7.35 & 24.85 \\
GRPO-RLVR    & \textbf{20.45} & 36.97 & 13.64 & 27.88 \\
\rowcolor[gray]{0.95} Ours & 17.12 & \textbf{47.88} & \textbf{14.09} & \textbf{33.94} \\
\bottomrule
\end{tabular}
\end{sc}
\end{small}
\vspace{-1mm}
\end{table}

\begin{table}[t]
\centering
\caption{Supplementary analysis of the GRPO training ceiling on Qwen2.5-7B-Instruct. We report Pass@1 (\%) and Pass@128 (\%) on WebShop and Sokoban. Compared with extended GRPO training, the on-policy RL+CD variant achieves higher performance, suggesting that interleaving Counterfactual Distillation with GRPO can further raise the training ceiling of outcome-driven RL.}
\label{tab:grpo_training_ceiling}
\vspace{2mm}
\begin{small}
\begin{sc}
\setlength{\tabcolsep}{5pt}
\renewcommand{\arraystretch}{0.95}
\begin{tabular}{@{}lcccc@{}}
\toprule
& \multicolumn{2}{c}{WebShop} & \multicolumn{2}{c}{Sokoban} \\
\cmidrule(lr){2-3} \cmidrule(lr){4-5}
Method & Pass@1 & Pass@128 & Pass@1 & Pass@128 \\
\midrule
GRPO (Main Table) & 67.45 & 85.40 & 58.15 & 68.00 \\
GRPO (+Steps) & 71.22 & 88.60 & 70.70 & 75.20 \\
\rowcolor[gray]{0.95} RL+CD(interleaved) & \textbf{74.33} & \textbf{91.00} & \textbf{78.70} & \textbf{81.20} \\
\bottomrule
\end{tabular}
\end{sc}
\end{small}
\vspace{-1mm}
\end{table}

\subsection{Capability Scaling: Pass@k Analysis}

\begin{figure*}[t]
  \centering
  \includegraphics[width=0.98\textwidth]{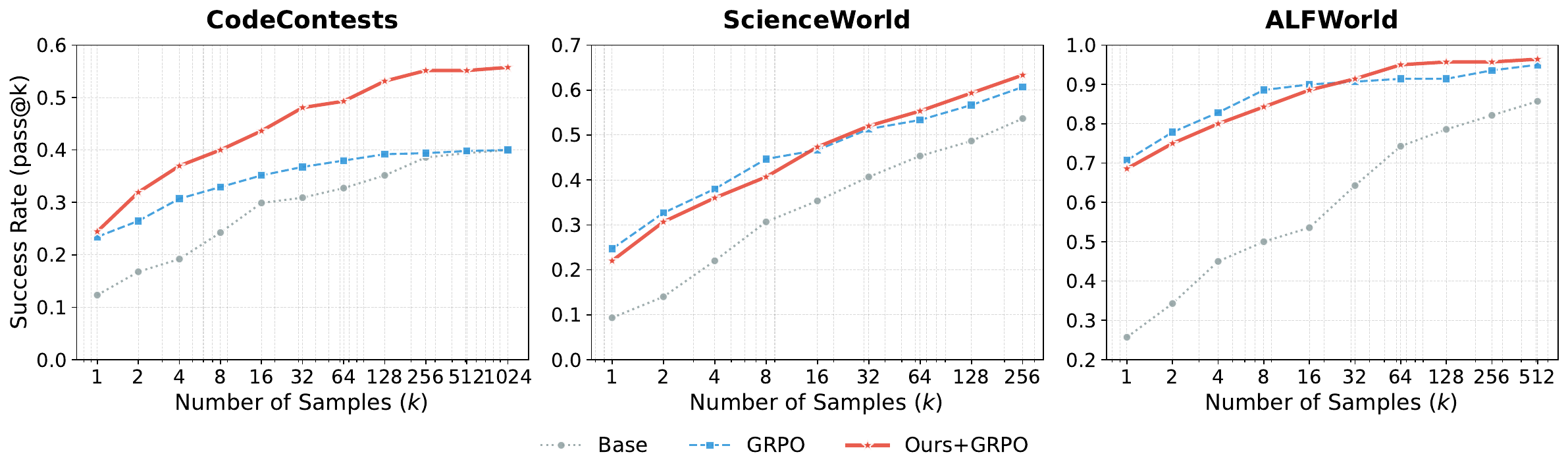}
  \vspace{-3mm}
  \caption{\textbf{Scaling results on different benchmarks.} We plot the Pass@$k$ success rate as a function of the number of samples $k$. Our method (red) consistently achieves higher efficiency and performance ceilings across all tasks compared to the baselines.}
  \vspace{-1mm}
  \label{fig:pass_k_scaling}
\end{figure*}

Figure~\ref{fig:pass_k_scaling} analyzes Pass@$k$ as the sampling budget increases, with task-specific caps of $k{=}512$ for ALFWorld, $k{=}256$ for ScienceWorld, and $k{=}1024$ for CodeContests. Across all benchmarks, our method consistently performs best in the large-$k$ regime, indicating a higher capability ceiling, with especially clear gains on CodeContests. The scaling curves further show that Counterfactual Distillation reaches comparable success rates with fewer samples and maintains stronger performance as $k$ grows. These results suggest that our method improves not only the attainable upper bound of the policy, but also the efficiency with which additional sampling translates into successful trajectories.

\subsection{Effectiveness Analysis}

\textbf{Tree-Structured Exploration for Experience Generation.}
Table~\ref{tab:vanilla_iterative_ours} compares three sampling strategies on CodeContests under the same execution budget: independent sampling, iterative refinement using previous execution feedback, and our tree-based rollback branching. Across all models, our Stage~1 strategy achieves the highest Pass@128, showing that feedback-guided branching from critical failure points is more effective than independent or linear refinement.

\vspace{0.3mm}
\textbf{Counterfactual Distillation.}
Table~\ref{tab:pass_scaling_ablation} compares rehearsal alone ($\mathcal{L}_{reh}$) with rehearsal plus counterfactual distillation ($\mathcal{L}_{cf}+\mathcal{L}_{reh}$) on ScienceWorld. While $\mathcal{L}_{reh}$ preserves successful behaviors through rejection-sampling-style imitation~\citep{ahn2024large}, adding $\mathcal{L}_{cf}$ substantially improves Pass@128 with comparable Pass@1 across model sizes, e.g., $59.33\%\rightarrow62.00\%$ on Qwen2.5-7B and $67.33\%\rightarrow72.00\%$ on Qwen2.5-14B. This indicates that counterfactual distillation is crucial for internalizing corrective revisions, while rehearsal mainly stabilizes training.

\begin{table}[t]
\centering
\caption{Ablation of counterfactual distillation on ScienceWorld. Adding $\mathcal{L}_{\mathrm{cf}}$ to rehearsal ($\mathcal{L}_{\mathrm{reh}}$) improves Pass@128 across all three backbones, with comparable Pass@1.}
\label{tab:pass_scaling_ablation}
\vspace{1mm}
\small
\setlength{\tabcolsep}{6pt}
\renewcommand{\arraystretch}{1.15}
\begin{tabular}{@{}lcccc@{}}
\toprule
\multirow{2}{*}{\textbf{Model}} &
\multicolumn{2}{c}{\textbf{Pass@1}} &
\multicolumn{2}{c}{\textbf{Pass@128}} \\
\cmidrule(lr){2-3}\cmidrule(lr){4-5}
& $\mathcal{L}_{\text{reh}}$
& $\mathcal{L}_{\text{cf}}+\mathcal{L}_{\text{reh}}$
& $\mathcal{L}_{\text{reh}}$
& $\mathcal{L}_{\text{cf}}+\mathcal{L}_{\text{reh}}$ \\
\midrule
Qwen2.5-7B   & 27.67 & 27.88 & 59.33 & 62.00 \\
Llama3.1-8B  & 22.54 & 22.70 & 57.33 & 59.33 \\
Qwen2.5-14B  & 37.17 & 36.50 & 67.33 & 72.00 \\
\bottomrule
\end{tabular}
\end{table}

\subsection{Ablation Study}

\begin{table}[t]
\centering
\caption{Ablation of the training context on Qwen2.5-7B-Instruct. Using the same revision data, CD (Ours) removes experience $e$ from the training input, while CD (keep exp) retains it. CD (Ours) achieves higher Pass@$128$ on both tasks with counterfactual supervision context.}
\label{tab:keep_exp_ablation}
\small
\setlength{\tabcolsep}{7pt}
\renewcommand{\arraystretch}{1.05}
\begin{tabular}{@{}lcccc@{}}
\toprule
& \multicolumn{2}{c}{WebShop} & \multicolumn{2}{c}{Sokoban} \\
\cmidrule(lr){2-3}\cmidrule(lr){4-5}
Method & Pass@$1$ & Pass@$128$ & Pass@$1$ & Pass@$128$ \\
\midrule
GRPO-RLVR & $\mathbf{67.32} \pm 1.17$ & 85.40 & $58.33 \pm 0.55$ & 68.00 \\
CD (keep exp) & $65.86 \pm 0.98$ & 84.80 & $60.23 \pm 0.65$ & 72.00 \\
CD (Ours) & $66.38 \pm 1.09$ & \textbf{89.20} & $\mathbf{65.32} \pm 0.98$ & \textbf{78.40} \\
\bottomrule
\end{tabular}
\end{table}

\noindent\begin{minipage}{\linewidth}
\textbf{Counterfactual Context Supervision.}
Table~\ref{tab:keep_exp_ablation} compares CD (Ours) with CD (keep exp) under matched training conditions (Appendix~\ref{app:keep_exp_protocol}).
CD (keep exp) reinforces revised actions with the hindsight experience $e$ that generated them.
CD (Ours) instead learns the same actions under the original context $(h_\tau,q)$ without $e$, achieving higher Pass@$128$ on both tasks.
These results suggest that counterfactual context supervision internalizes hindsight-guided action improvements in model parameters,
enabling effective decisions without experience at test time.
\end{minipage}

\vspace{0.3mm}
\textbf{Scaling Behavior Across Model Sizes.} As shown in Fig.~\ref{fig:scale_model}, Pass@128 exhibits a consistent scaling trend across benchmarks. Increasing the model size from Qwen2.5-7B to 14B yields gains of +0.6~pp on WebShop and +10.0~pp on ScienceWorld, and the transition from Qwen2.5-32B to 72B yields +13.83~pp on CodeContests. Our method consistently outperforms all baselines across these scales. The robustness of these gains across both Qwen and Llama architectures demonstrates that Counterfactual Distillation effectively leverages increased model capacity to raise the performance ceiling.

\vspace{0.3mm}
\textbf{Out-of-Distribution Generalization.} To evaluate Out-of-Distribution (OOD) generalization, we test models trained on CodeContests on the held-out MBPP benchmark. As shown in Table~\ref{tab:ood_mbpp_p128}, our method shows superior robustness to distribution shift than the GRPO baseline. While GRPO leads to severe degradation on OOD tasks (e.g., a drop of 4.2 percentage points on Llama3-70B), Counterfactual Distillation not only mitigates this drop but still surpasses the original base model. This suggests that our method helps the model learn fundamental reflective agency rather than overfitting to dataset-specific shortcuts.

We provide a more detailed ablation study in Appendix~\ref{app:ablation-study}.
\section{Related Work}\label{related_work}

\textbf{Self-Evolving LLM Agents.}
A growing body of work studies how LLM agents improve through interaction, reflection, and accumulated experience.
One line externalizes learning into prompts, memories, or reusable experience libraries without directly updating model weights, as in ReasoningBank~\citep{ouyang2025reasoningbank}, FLEX~\citep{cai2025flex}, and ACE~\citep{zhang2025agentic}.
Another line performs \emph{training-based} self-evolution by iteratively turning interaction data into better policies, skills, or strategic abstractions: SKILLRL~\citep{xia2026skillrl} and EvolveR~\citep{wu2025evolver} distill trajectories into reusable skills or principles, while Agent0~\citep{xia2025agent0}, Absolute Zero~\citep{zhao2025absolute}, and CodeIt~\citep{butt2024codeit} couple self-improvement with automatic task generation, self-play style curricula, or hindsight relabeling.
A third line focuses on \emph{test-time} self-improvement, either through lightweight test-time fine-tuning on synthesized data~\citep{acikgoz2025self} or through repeated evolution of prompts, memory, and tool-use configurations across attempts~\citep{he2025evotest}.
Our work emphasizes internalizing feedback-grounded exploratory experience into the policy for long-horizon interaction.

\textbf{Learning from Experience.}
Learning from experience is a long-standing goal in agentic systems.
Before LLMs, model-based RL such as World Models~\citep{ha2018world} and Dreamer~\citep{hafner2023mastering,hafner2020mastering} learned latent dynamics to imagine future outcomes for planning.
For LLM agents, Early Experience (EarlyExp)~\citep{zhang2025agent} learns from self-generated traces via implicit world modeling and self-reflection; Agent Q~\citep{putta2024agent} combines guided MCTS and self-critique with off-policy DPO to learn from successes and failures; and SPA~\citep{chen2025internalizing} internalizes world models via self-play finetuning.
Related rollback or hindsight-based methods further exploit failed trajectories: HOPE~\citep{wangroad} uses hindsight counterfactual actions to drive multi-turn RL exploration, while GA-Rollback~\citep{li2025generator} triggers stepwise rollback to prevent error propagation.
In contrast, Counterfactual Distillation explicitly localizes critical failure points during tree-structured interaction and distills the resulting experience-guided corrective actions under the original decision contexts without the experience text, thereby internalizing the revisions into a single-rollout policy without requiring external memory or heavy test-time search.

Additional background on LLM agents, RLVR, and Pass@$k$ optimization is provided in Appendix~\ref{app:related-work}.

\section{Limitations and Conclusion}

We presented Counterfactual Distillation, a framework that internalizes reflective experience into LLM agents by converting rollout-time feedback into counterfactual policy supervision. Through tree-structured exploration, the agent localizes failure points, generates experience-guided revisions, and distills these revisions under the original decision contexts. Across long-horizon agentic benchmarks, this improves Pass@$k$, especially at large $k$, indicating broader behavioral coverage and a higher capability ceiling than outcome-driven RLVR.

The main limitation is its reliance on informative feedback and replayable environment states. When feedback is sparse, delayed, or ambiguous, reflected experience may be unreliable; when environments are highly stochastic or stateful, exact rollback may be impractical. Extending the method to such settings may require additional state tracking or approximate replay mechanisms.

\section*{Acknowledgements}
We sincerely thank the anonymous reviewers for their insightful and constructive feedback.
We gratefully acknowledge MBZUAI and NVIDIA for providing the computational resources that made this work possible.

\newpage
\bibliography{main}
\bibliographystyle{plainnat}

\newpage
\appendix

\onecolumn

\section{Algorithm Detail}
\begin{algorithm}[h]
  \caption{\textsc{Counterfactual Distillation} Stage~1: Tree-Structured Experience Search}
  \label{alg:lefe_generation}
  \begin{algorithmic}[1]
    \STATE {\bfseries Input:} environment $\mathcal{E}$, policy $\pi_\theta$, instruction $q$, period $K$, max steps $T$, budget $B$
    \STATE Initialize $\mathcal{W} \leftarrow \{(1, \emptyset, \emptyset)\}$; \quad $\mathcal{D}_{\text{traj}} \leftarrow \emptyset$; \quad $\mathcal{D}_{\text{refl}} \leftarrow \emptyset$
    \STATE $count \leftarrow 0$
    \WHILE{$\mathcal{W} \neq \emptyset$ {\bfseries and} $count < B$}
      \STATE Dequeue $(\tau, e, a_{1:\tau-1})$ from $\mathcal{W}$; \quad $count \leftarrow count + 1$
      \STATE $\mathrm{Reset}(\mathcal{E})$; \quad $\mathrm{Replay}(\mathcal{E}, a_{1:\tau-1})$ \COMMENT{Restore $E_\tau$ and $h_\tau$}
      \STATE $t \leftarrow \tau$
      \WHILE{$t \leq T$ {\bfseries and} $\mathcal{E}$ not terminal}
        \IF{$t = \tau$ {\bfseries and} $e \neq \emptyset$}
          \STATE $a_t \sim \pi_\theta(\cdot \mid h_t, q, e)$ \COMMENT{Generate improved action $a_\tau'$}
          \STATE $\mathcal{D}_{\text{refl}} \leftarrow \mathcal{D}_{\text{refl}} \cup \{(h_t, e, a_t)\}$ \COMMENT{Collect experience-guided sample}
        \ELSE
          \STATE $a_t \sim \pi_\theta(\cdot \mid h_t, q)$
        \ENDIF
        \STATE $E_{t+1}, o_{t+1} \leftarrow \mathrm{Step}(E_t, a_t)$; \quad $h_{t+1} \leftarrow h_t \cup (a_t, o_{t+1})$
        
        \IF{$t \bmod K = 0$ {\bfseries or} $\mathcal{E}$ fails}
          \STATE $(\hat{\tau}, \hat{e}) \sim \pi_\theta(\cdot \mid h_{t+1}, p_{\text{refl}})$ \COMMENT{Identify failure and provide guidance}
          \STATE Enqueue $(\hat{\tau}, \hat{e}, a_{1:\hat{\tau}-1})$ into $\mathcal{W}$
        \ENDIF
        
        \IF{$\mathcal{E}$ succeeds} 
          \STATE $\mathcal{D}_{\text{traj}} \leftarrow \mathcal{D}_{\text{traj}} \cup \{h_{t+1}\}$ 
        \ENDIF
        \STATE $t \leftarrow t + 1$
      \ENDWHILE
    \ENDWHILE
  \end{algorithmic}
\end{algorithm}

\section{Related Work}
\label{app:related-work}

\textbf{LLM-based Agents.} Large Language Models (LLMs) have evolved from passive generators into autonomous agents capable of tool-use and multi-step reasoning. Early work largely relies on prompting to elicit agent behaviors without weight updates, including ReAct~\citep{yao2022react}, Reflexion~\citep{shinn2023reflexion}, and Tree of Thoughts~\citep{yao2023tree}. Recently, reinforcement-learning (often RLVR) has become a major route to improve reasoning robustness and long-horizon decision making, exemplified by DeepSeek-R1~\citep{guo2025deepseek} and OpenAI o1~\citep{jaech2024openai}. Building on this direction, systems such as verl-agent/GiGPO~\citep{feng2025group}, rLLM~\citep{rllm2025}, and Agent-R1~\citep{cheng2025agent} scale multi-turn RL for tool-using agents, while open-weight releases (e.g., Kimi K2~\citep{team2025kimi}, GLM-4.5/4.7~\citep{zeng2025glm,zai2025glm47}, MiniMax-M2~\citep{minimax2025m2agent}) further accelerate reproducible agent research.

\textbf{Reinforcement Learning with Verifiable Rewards (RLVR).} RLVR post-trains LLMs with automatically verifiable signals (e.g., exact-match answers, unit tests, theorem provers, or environment states), and has become a widely adopted recipe for improving reasoning on mathematics and programming-style tasks~\citep{guo2025deepseek,yu2025dapo}. PPO~\citep{schulman2017proximal} is a standard baseline, while critic-free group-based optimizers such as GRPO~\citep{shao2024deepseekmath,guo2025deepseek} and step-aware variants like GiGPO~\citep{feng2025group} improve stability and credit assignment; these recipes have been reproduced and scaled in open settings (e.g., SimpleRL-Zoo~\citep{zeng2025simplerl}, ORZ~\citep{hu2025open}, and Skywork-OR1~\citep{he2025skywork}). Beyond sparse outcome rewards, recent work explores denser step-wise supervision via process reward modeling and automated process verifiers~\citep{setlur2024rewarding} as well as implicit/dense reward construction~\citep{cui2025process}.
RLVR is also being scaled and extended through efficient small-model RL scaling~\citep{deepscaler2025}, budget-aware optimization~\citep{qi2025optimizing}, and system-level acceleration (e.g., asynchronous rollouts)~\citep{fu2025areal}, and has expanded to new verifiable domains such as Verilog generation~\citep{zhu2025qimeng} and Lean-based theorem proving~\citep{ji2025leanabell,shang2025stepfun}.

\textbf{Pass@k Optimization Methods.}
Recent work has studied Pass@k not only as an evaluation metric, but also as an objective or diagnostic for balancing exploration and exploitation in reasoning models.
Several studies observe that RLVR can improve Pass@1 while providing limited gains, or even causing degradation, at large $k$, suggesting that outcome-driven training may sharpen existing solution modes rather than expand the model's capability boundary~\citep{yue2025does,dang2025weight}.
To address this, Pass@k Training~\citep{chen2025pass} directly optimizes a Pass@k-oriented objective to adaptively balance exploration and exploitation, while Pass@k Policy Optimization~\citep{walder2025pass} formulates policy optimization around the utility of multiple sampled outputs.
Other works improve large-$k$ behavior by encouraging broader exploration or mitigating diversity collapse, such as outcome-based exploration~\citep{song2025outcome}, divergence choices that better preserve policy support~\citep{li2025choice}, differential smoothing~\citep{gai2025differential}, and systematic analyses of exploration mechanisms in RLVR~\citep{deng2025trial}.
Unlike these methods, which mainly focus on final-answer diversity or sample-set optimization, Counterfactual Distillation targets long-horizon interactive agents: it uses environment feedback to identify critical failure points and distills experience-guided corrective decisions into the policy, thereby improving Pass@k through decision-level behavioral coverage.

\section{Ablation Study Results}
\label{app:ablation-study}

\begin{figure}[t]
\vspace{-1mm}
\centering
\includegraphics[width=0.6\linewidth]{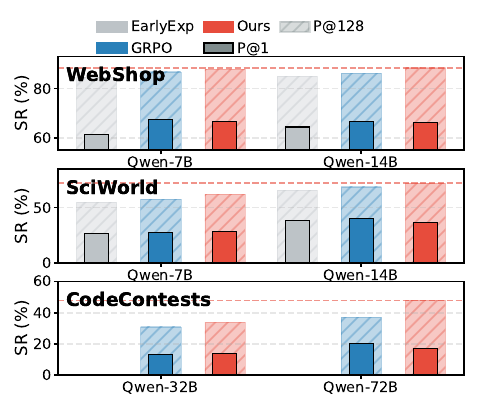}
\vspace{-2mm}
\caption{Main results on WebShop and SciWorld. Bars represent \textbf{Pass@1} (solid) and \textbf{Pass@k} (hatched) (\%). \textbf{Ours} consistently outperforms GRPO and other baselines across different model architectures and scales.}
\label{fig:scale_model}
\vspace{-3.5mm}
\end{figure}

\begin{table}[t]
\centering
\caption{OOD generalization on MBPP after training on CodeContests. Values are Pass@128 (\%); superscripts show absolute changes from the base model, with red for drops and blue for gains. Counterfactual Distillation preserves OOD performance better than GRPO and can surpass the base model.}
\label{tab:distill_ood}
\label{tab:ood_mbpp_p128}
\vspace{1mm}
\small
\setlength{\tabcolsep}{4pt}
\renewcommand{\arraystretch}{1.15}
\begin{tabular}{@{}lccc@{}}
\toprule
\textbf{Method}
& \textbf{Qwen2.5-32B}
& \textbf{Qwen2.5-72B}
& \textbf{Llama3-70B} \\
\midrule
Base
& 85.45
& 83.33
& 78.31 \\
GRPO
& 81.22$^{\textcolor{red}{-4.2}}$
& 81.22$^{\textcolor{red}{-2.1}}$
& 74.07$^{\textcolor{red}{-4.2}}$ \\
\midrule
\textbf{Counterfactual Distillation}
& \textbf{85.45}$^{\textcolor{blue}{+0.0}}$
& \textbf{85.13}$^{\textcolor{blue}{+1.8}}$
& \textbf{79.63}$^{\textcolor{blue}{+1.3}}$ \\
\bottomrule
\end{tabular}
\end{table}

\begin{table}[t]
\centering
\caption{CodeContests Pass@128 under different sampling strategies with Independent Sampling (IS), Iterative Refinement (IR), and our stage~1.}
\label{tab:vanilla_iterative_ours}
\vspace{0.6mm}
\small
\setlength{\tabcolsep}{6pt}
\renewcommand{\arraystretch}{1.15}
\begin{tabular}{lccc}
\toprule
\textbf{Model} & \textbf{IS} & \textbf{IR} & \textbf{Ours (Stage~1)} \\
\midrule
Qwen2.5-32B & 48.92 & 51.48 & \textbf{55.52} \\
Qwen2.5-72B & 48.65 & 49.52 & \textbf{54.30} \\
Llama3-70B  & 30.20 & 38.10 & \textbf{42.50} \\
\bottomrule
\end{tabular}
\vspace{-3mm}
\end{table}

\begin{table}[!t]
\centering
\caption{Ablation study on the impact of auxiliary training stages (i.e., EarlyExp (EE) and GRPO (RL)). For ALFWorld, Ours is GRPO$\rightarrow$CD, and w/ EE adds EE before GRPO. For CodeContests, Ours is CD, and w/ RL adds GRPO before CD. Results are reported as Pass@$1$ / Pass@$128$ (\%). While these additional stages consistently enhance single-rollout performance (Pass@$1$), they may occasionally constrain the exploration ceiling (Pass@$128$).}
\vspace{0.6mm}
\label{tab:synergy}
\resizebox{0.6\linewidth}{!}{
\scriptsize
\setlength{\tabcolsep}{5pt}
\begin{tabular}{@{}ll cc@{}}
\toprule
\textbf{Task} & \textbf{Model} & \textbf{Ours} & \textbf{Ours + Enhancement} \\ 
\midrule
\multirow{2}{*}{\textit{ALFWorld}} 
& Qwen2.5-7B  & 67.50 / 94.29 & \textbf{67.86} / \textbf{95.71} (w/ EE) \\
& Llama3.1-8B & 71.79 / \textbf{96.43} & \textbf{74.46} / 93.57 (w/ EE) \\
\midrule
\multirow{3}{*}{\textit{CodeContests}} 
& Qwen2.5-32B & 9.34 / \textbf{34.35}  & \textbf{15.38} / 34.05 (w/ RL) \\
& Qwen2.5-72B & 11.15 / 40.00 & \textbf{17.12} / \textbf{47.88} (w/ RL) \\
& Llama3-70B & 9.33 / 30.30 & \textbf{14.09} / \textbf{33.94} (w/ RL) \\
\bottomrule
\end{tabular}
}
\vspace{-2mm}
\end{table}

\begin{table}[t]
\centering
\caption{Comparison with Early Experience-SR on ScienceWorld using Qwen2.5-7B-Instruct. We report Pass@1 (\%) and Pass@128 (\%). Counterfactual Distillation achieves the best Pass@128, indicating stronger exploration capability under large sampling budgets.}
\label{tab:earlyexp_sr_comparison}
\vspace{2mm}
\begin{small}
\begin{sc}
\setlength{\tabcolsep}{6pt}
\renewcommand{\arraystretch}{0.95}
\begin{tabular}{@{}lcc@{}}
\toprule
Method & Pass@1 & Pass@128 \\
\midrule
Base & 7.00 & 47.33 \\
Imitation Learning & 37.33 & 80.67 \\
EarlyExperience-SR & \textbf{40.87} & 82.67 \\
\rowcolor[gray]{0.95} Counterfactual Distillation & 40.50 & \textbf{86.00} \\
\bottomrule
\end{tabular}
\end{sc}
\end{small}
\vspace{-1mm}
\end{table}

\paragraph{Synergy and Trade-offs with Auxiliary Training Target.} We investigated the synergy of Counterfactual Distillation with EarlyExp and GRPO. For agentic tasks, GRPO first builds basic interaction skills; coding models already have basic coding skills, so we also evaluate CD alone on CodeContests. CodeContests does not use EE. As presented in Table~\ref{tab:synergy}, integrating these techniques consistently improves Pass@$1$ success rates across all evaluated benchmarks: For instance, adding EarlyExp before GRPO for Llama3.1-8B on ALFWorld increases accuracy from 71.79\% to 74.46\%, while adding GRPO before CD boosts Qwen2.5-72B on CodeContests from 11.15\% to 17.12\%. However, these gains in exploitation do not always translate to enhanced model capability at a higher sampling budget. We observe that Pass@$128$ performance can sometimes decline with these additional training stages, notably for Llama3.1-8B (96.43\% $\rightarrow$ 93.57\%) and Qwen2.5-32B (34.35\% $\rightarrow$ 34.05\%), suggesting that while EarlyExp and GRPO effectively align the primary training target, they may narrow the exploration space. In contrast, Counterfactual Distillation consistently improves the model's capacity as analyzed in \S\ref{sec:main-result}. 

\paragraph{Comparison with Early Experience-SR.}
Early Experience is an important baseline for learning from agent interaction. In the main experiments, we primarily compare with its Implicit World Modeling (IWM) variant, since the Self-Reflection (SR) variant relies on expert demonstrations and is therefore less aligned with our main comparison setting, where supervision is obtained from model-generated interaction and environment feedback. To provide a more complete comparison, we additionally reproduce Early Experience-SR on ScienceWorld with Qwen2.5-7B-Instruct. Following the SR setup, we first train an Imitation Learning baseline using expert demonstrations, and then train Early Experience-SR with a mixed IL+SR dataset. In our reproduction, the Imitation Learning data scale is about 14k samples, the mixed IL+SR training set is about 28k samples, and the Counterfactual Distillation training data contains about 4k samples. As shown in Table~\ref{tab:earlyexp_sr_comparison}, EarlyExperience-SR achieves the best Pass@1 score of 40.87\%, slightly higher than Counterfactual Distillation at 40.50\%. However, Counterfactual Distillation achieves a substantially higher Pass@128 score, improving from 82.67\% to 86.00\%. This result suggests that Counterfactual Distillation more effectively expands the policy's exploration space and achieves a higher capability ceiling under large sampling budgets.

\paragraph{Intermediate Pass@$k$ Results.}
Table~\ref{tab:keep_exp_scaling} extends the experience-removal ablation with intermediate sampling budgets on WebShop and Sokoban.
The benefit is clearest at larger sampling budgets; CD does not lead at every intermediate $k$ (e.g., WebShop at $k=8$).

\begin{table}[t]
\centering
\caption{Intermediate Pass@$k$ results (\%) on Qwen2.5-7B-Instruct under independent sampling. These point estimates extend the Pass@$128$ results in Table~\ref{tab:keep_exp_ablation}.}
\label{tab:keep_exp_scaling}
\small
\setlength{\tabcolsep}{6pt}
\renewcommand{\arraystretch}{1.05}
\begin{tabular}{@{}llccccc@{}}
\toprule
Task & Method & $k=8$ & $k=16$ & $k=32$ & $k=64$ & $k=128$ \\
\midrule
\multirow{3}{*}{WebShop} & GRPO-RLVR & 76.20 & 77.80 & 82.00 & 84.20 & 85.40 \\
& CD (keep exp) & \textbf{76.60} & 78.40 & 81.20 & 83.40 & 84.80 \\
& CD & 75.80 & \textbf{80.80} & \textbf{84.40} & \textbf{86.40} & \textbf{89.20} \\
\midrule
\multirow{3}{*}{Sokoban} & GRPO-RLVR & 63.20 & 64.60 & 66.00 & 67.40 & 68.00 \\
& CD (keep exp) & 69.20 & 70.40 & 71.20 & 71.80 & 72.00 \\
& CD & \textbf{75.20} & \textbf{77.40} & \textbf{78.00} & \textbf{78.40} & \textbf{78.40} \\
\bottomrule
\end{tabular}
\end{table}

\section{Implementation Details}
\label{implementation-detals}

\subsection{Summary}
Table~\ref{tab:data_stats} summarizes the data scale used across all benchmarks.
For each dataset, we report the exact number of instances in the task split (training set and test set).
In addition, we list the average amount of auxiliary SFT data (EarlyExp and our Counterfactual Distillation) distilled from rollouts; these values are approximate and meant to indicate the typical order of magnitude (in thousands) rather than an exact count, since the final number can vary slightly across runs.

\begin{table}[!ht]
\vspace{-1mm}
\centering
\caption{Dataset statistics and approximate numbers of training samples used by auxiliary supervision.
Training/test set sizes are exact instance counts, while EarlyExp and Counterfactual Distillation refer to the average scale (order of magnitude) of distilled samples used for SFT across runs.}
\label{tab:data_stats}
\vspace{1mm}
\small
\setlength{\tabcolsep}{7pt}
\renewcommand{\arraystretch}{1.25}
\begin{tabular}{@{}lcccc@{}}
\toprule
\textbf{Dataset} & \textbf{\#Training Set} & \textbf{\#Test Set} & \textbf{\#EarlyExp samples} & \textbf{\#Counterfactual Distillation samples} \\
\midrule
ALFWorld     & 3553 & 140 & 10k & 12k \\
ScienceWorld & 2000 & 150 & 6k  & 12k \\
WebShop      & 5000 & 500 & 15k & 10k \\
Sokoban      & 5000 & 500 & 10k & 10k \\
CodeContests & 5000 & 165 & N/A   & 10k \\
\bottomrule
\end{tabular}
\end{table}

\subsection{Computational Resources}

All experiments are conducted on NVIDIA GB200 nodes. 
Each node contains 4 GPUs. 
For the GRPO training on ScienceWorld and the large-model experiments on CodeContests with 70B-scale models, we use 2 nodes, corresponding to 8 GPUs in total. 
All other experiments are conducted on 1 node with 4 GPUs.

Table~\ref{tab:compute_cost} summarizes the approximate computational cost of different training or data-construction procedures. 
For GRPO, the runtime varies substantially across datasets: ScienceWorld takes about 20 hours, CodeContests takes about 15 hours, and the remaining datasets are generally completed within 10 hours. 
For Early Experience and our method, the reported time includes both data collection and supervised training. 
For ACE, the reported time mainly corresponds to collecting interaction data and constructing the knowledge base. 
The number of steps denotes the approximate amount of data used in each procedure, measured in GRPO-equivalent training steps.

\begin{table}[!ht]
\vspace{-1mm}
\centering
\caption{Approximate computational cost of different methods. 
The runtime is reported in hours. 
The number of steps denotes the approximate amount of data used by each method, measured in GRPO-equivalent training steps.}
\label{tab:compute_cost}
\vspace{1mm}
\small
\setlength{\tabcolsep}{12pt}
\renewcommand{\arraystretch}{1.25}
\begin{tabular}{@{}lcccc@{}}
\toprule
\textbf{} & \textbf{GRPO} & \textbf{Early Experience} & \textbf{ACE} & \textbf{Ours} \\
\midrule
Time (hr) & 6--20 & 2 & 3 & 3 \\
Steps     & 100--150 & 10 & 5 & 10 \\
\bottomrule
\end{tabular}
\end{table}

\subsection{Implementation of Environment}

\paragraph{ALFWorld.}
ALFWorld is a household embodied task suite built on text-based environments.
At each step, the environment executes the agent's action and returns a textual observation describing the execution result and the updated scene state.
In addition, ALFWorld provides a list of \emph{admissible actions} for the next step, which we expose to the model as action candidates.

\paragraph{ScienceWorld.}
ScienceWorld is an interactive scientific reasoning environment in which the agent solves multi-step tasks by manipulating objects and instruments.
After executing an action, the environment returns a textual observation describing the action outcome and the resulting world state.
Unlike ALFWorld, ScienceWorld exposes a set of currently \emph{available objects}.
The agent must compose valid actions by selecting an action template and instantiating it with objects from this list (e.g., \texttt{use [object] on [object]}), following the environment's action syntax.

\paragraph{WebShop.}
WebShop simulates goal-directed online shopping as a browser-like interaction loop.
At each step, the environment renders the current webpage content (including product lists, descriptions, and navigation elements).
The agent can either issue a \emph{search} action by providing a keyword query, or perform a \emph{click} action by selecting a clickable element (e.g., a product entry, category link, or a back button) to navigate across pages toward the target purchase.

\paragraph{Sokoban.}
Sokoban is a grid-based puzzle environment in which the agent must push boxes onto designated target locations through sequential movement decisions.
At each step, the environment executes one of four directional actions (\emph{up}, \emph{down}, \emph{left}, \emph{right}) and returns the updated grid state.
In our setup, we use a \emph{text-mode} representation of the environment, where each grid cell is rendered as a symbolic token: walls, floor tiles, targets, boxes, the player are represented by symbols such as \texttt{\#}, \texttt{\_}, \texttt{O}, \texttt{X}, \texttt{P}.
This textual grid is exposed to the model as the observation, and the agent must infer spatial structure, object interactions, and long-horizon box-pushing plans directly from these symbolic layouts.

\paragraph{CodeContests.}
CodeContests is a competitive programming benchmark where each episode corresponds to a single coding problem.
In our setup, at every round, the model produces a \emph{complete} candidate solution (full code) for the problem.
The environment compiles and executes the submitted code against test cases and returns execution feedback.
On the public test set, the environment provides concrete runtime outputs (e.g., failing case outputs, errors, or mismatched results), whereas on the private set it reports only an aggregate correctness score.
The model uses this feedback to generate a revised full-program submission in the next round.

\subsection{Implementation of Our Method  Counterfactual Distillation}

\paragraph{Stage 1: Tree-Structured Exploration for Experience Generation.}
During rollout, we trigger \emph{rollback reflection} either (i) every fixed interval of $K$ interaction rounds, or (ii) immediately when an episode terminates in failure. 
At each reflection, we provide the model with the full interaction history up to the current step and prompt it to (a) select a specific \emph{suboptimal round} $\tau$ to roll back to, and (b) summarize a concise \emph{experience} $e$ (diagnosis-and-fix instruction) grounded in the observed feedback.
The chosen $\tau$ is intended to correspond to the decision that caused the trajectory to enter an incorrect branch or become unproductive.
A concrete prompt example is shown in Appendix~\ref{Counterfactual Distillation-prompt}.
In our implementation, we set $K{=}10$ for ALFWorld and ScienceWorld, $K{=}5$ for WebShop and Sokoban, and $K{=}4$ for CodeContests.
Notably, CodeContests has a maximum horizon of $4$, so rollback is only invoked when the attempt budget is exhausted.

\paragraph{Stage 2: Counterfactual Distillation.}
We construct two types of supervised data for standard next-token likelihood training. 
First, \emph{Behavior Rehearsal} samples successful trajectories to preserve the agent's base competence.
We randomly select $20\%$ of successful rollouts to form the rehearsal set $\mathcal{D}_{\mathrm{reh}}$.
Second, \emph{Experience-to-Policy Distillation} builds a \emph{counterfactual} dataset $\mathcal{D}_{\mathrm{cf}}$ from branching points: when an experience $e$ injected at round $\tau$ yields an improved action $a'_{\tau}$, we treat $a'_{\tau}$ as the counterfactual target for the \emph{original} history without providing $e$ at training time.
In practice, we sample $3$ counterfactual (branching) instances per task to populate $\mathcal{D}_{\mathrm{cf}}$.
We then perform SFT on the union of $\mathcal{D}_{\mathrm{reh}}$ and $\mathcal{D}_{\mathrm{cf}}$ with batch size $128$ and learning rate $1{\times}10^{-6}$ for $2$--$3$ epochs, depending on the dataset and model scale.

\paragraph{Protocol for the Experience-Removal Ablation.}
\label{app:keep_exp_protocol}
Both CD variants start from the same GRPO-trained Qwen2.5-7B-Instruct checkpoint and use the same reflection-generated revision data, training budget, and evaluation protocol.
CD (keep exp) predicts $a'_\tau$ from $(h_\tau,q,e)$, whereas CD predicts the same target from $(h_\tau,q)$; both are evaluated without $e$, rollback, or experience-guided branching.
The Pass@$1$ values in Table~\ref{tab:keep_exp_ablation} are mean $\pm$ SD over 32 complete stochastic evaluations of each fixed checkpoint, with one rollout per test instance in each evaluation.
These are repeated evaluations, not independent training runs, and need not equal the original point estimates in Table~\ref{tab:main_results}.
The Pass@$128$ values in Table~\ref{tab:keep_exp_ablation} are point estimates under independent sampling; the supplementary variability estimates are reported separately in Appendix~\ref{app:eval_variability}.
Table~\ref{tab:keep_exp_scaling} reports the intermediate Pass@$k$ results.

\paragraph{Rollback-Tree Hyperparameters}

We use a small set of task-dependent hyperparameters for rollback-tree exploration. 
Across all main experiments, we set the maximum rollback-tree depth to $d=3$. 
The reflection frequency $f$ is chosen according to the typical interaction horizon of each environment: 
$f=10$ for ALFWorld and ScienceWorld, $f=5$ for WebShop, and $f=4$ for CodeContests. 
Intuitively, environments with longer horizons can tolerate less frequent reflection, while shorter-horizon or faster-feedback tasks benefit from more frequent opportunities to summarize experience and branch from earlier decisions.

To examine whether our method is sensitive to these choices, we conduct a small hyperparameter ablation on ScienceWorld. 
As shown in Table~\ref{tab:rollback_hyperparam_ablation}, increasing the tree depth from $d=3$ to $d=5$ yields a small improvement, while making reflection overly frequent ($f=5$) slightly hurts performance. 
This is likely because frequent reflection consumes the limited rollout budget more quickly, leaving fewer effective attempts for completing the task. 
Overall, the results suggest that our method is not highly sensitive to the rollback-tree settings within a reasonable range.

\begin{table}[!ht]
\vspace{-1mm}
\centering
\caption{Hyperparameter ablation of rollback-tree depth $d$ and reflection frequency $f$ on ScienceWorld(Based on imitation learning). 
The results suggest that our method is not highly sensitive to these settings: larger depth gives a small gain, while overly frequent reflection slightly degrades performance.}
\label{tab:rollback_hyperparam_ablation}
\vspace{1mm}
\small
\setlength{\tabcolsep}{12pt}
\renewcommand{\arraystretch}{1.25}
\begin{tabular}{@{}lccc@{}}
\toprule
\textbf{Metric} & \textbf{$d=3, f=10$} & \textbf{$d=5, f=10$} & \textbf{$d=3, f=5$} \\
\midrule
Pass@1   & 38.83 & 40.50 & 38.00 \\
Pass@128 & 85.33 & 86.00 & 83.33 \\
\bottomrule
\end{tabular}
\end{table}
\subsection{Implementation of Baselines}

\subsubsection{GRPO-RLVR}

\paragraph{Interactive agentic benchmarks (ALFWorld, ScienceWorld, WebShop, Sokoban).}
For ALFWorld, ScienceWorld, WebShop, and Sokoban, we train the GRPO-RLVR baseline using the \textsc{Verl-Agent} framework.
ALFWorld, WebShop, and Sokoban are officially supported in the repository, and we directly follow the provided environment setup, default training configurations, and reward definitions.
ScienceWorld is not officially implemented in \textsc{Verl-Agent}; we therefore implement a compatible environment wrapper following the official interfaces and guidelines, while keeping the observation/action formatting and data structures consistent with existing tasks in the framework.
Unless stated otherwise, we use the same default training setup and reward scheme as in \textsc{Verl-Agent}.
The only ScienceWorld-specific modification is in the memory module: we keep at most the most recent $10$ interaction turns in the context.

\paragraph{Competitive programming benchmark (CodeContests).}
For CodeContests, we train GRPO-RLVR using the \textsc{Verl} framework.
We follow the environment setup and evaluation protocol of CTRL: a solution receives reward of $1$ if and only if the generated program passes \emph{all} test cases, including both public and private tests; otherwise the reward is $0$.
We set the GRPO learning rate to $1\times 10^{-6}$, the training batch size to $128$, and use $n=8$ rollouts per prompt.
We train for $3$ epochs on the CodeContests training set.

\subsubsection{Early Experience}
In this section, we describe our reproduction of the Early Experience baseline, with a focus on its key component, \emph{Implicit World Modeling} (IWM).
In short, IWM trains the agent to \emph{predict the next environment observation} conditioned on the current interaction context and a candidate action, thereby implicitly capturing environment dynamics via supervised next-token prediction.

EarlyExp constructs IWM supervision using expert demonstrations. In our setting, we focus on interaction-driven learning: both our method and the GRPO-RLVR baseline obtain supervision from environment feedback rather than ground-truth trajectories. Therefore, for a fair comparison, we implement IWM supervision purely from model-generated rollouts:
we rollout the \emph{base model} on the training set and collect action--observation pairs, using the environment-returned observation as the prediction target.
For each task instance, we perform $4$ independent rollouts, select the highest-scoring attempt, and uniformly sample $3$ action--observation pairs from that attempt to form the IWM supervision set. We then perform SFT on it with batch size 64 and learning rate $1{\times}10^{-6}$ for 1 epoch.
An example prompt is shown in \ref{ee-prompt}.

\subsubsection{ACE}
ACE proposes a prompt-based self-evolving agent framework in which experience is accumulated in an external \emph{playbook} rather than internalized through parameter updates.
The playbook is a structured context memory that stores reusable task-solving strategies, heuristics, and reflective insights extracted from past interactions.
Its implementation is organized around three main roles: a \emph{Generator} that interacts with the environment to solve tasks, a \emph{Reflector} that analyzes trajectories and summarizes useful takeaways, and a \emph{Curator} that updates the playbook by incorporating new entries.

The open-source implementation of ACE is built on AppWorld, which serves as a representative interactive agent benchmark.
In our experiments, we extend the official implementation to ALFWorld, WebShop, ScienceWorld, and Sokoban by following the same overall ACE pipeline.
Specifically, we reuse the official prompts for the Generator, Reflector, and Curator roles, while keeping the task execution environments and interaction formatting consistent with the \textsc{Verl-Agent} setup used in our other interactive benchmarks.

Since the current ACE implementation only appends new playbook entries and does not perform effective pruning or consolidation, the playbook can grow quickly and lead to excessive context length.
To control context size, we cap the number of playbook entries during training at $100$.
At test time, we use the playbook obtained after training as part of the model context for downstream evaluation.

\newpage
\subsection{Evaluation Protocol and Variability}
\label{app:eval_variability}
We evaluate Qwen2.5-7B-Instruct on the four agent benchmarks using the same test sets, decoding settings, and success criteria as in the main experiments.
All rollouts are sampled directly from fixed checkpoints, without Stage~1 rollback or experience-guided branching.

\paragraph{Pass@$1$.}
We perform 32 complete independent stochastic evaluations, with exactly one rollout per test instance in each evaluation.
Table~\ref{tab:pass1_variability} reports the mean and sample SD of the resulting dataset-level success rates.

\begin{table}[!ht]
\centering
\caption{Pass@$1$ variability on Qwen2.5-7B-Instruct. We report mean $\pm$ SD (\%) over 32 complete stochastic evaluations of each fixed checkpoint.}
\label{tab:pass1_variability}
\vspace{1mm}
\small
\setlength{\tabcolsep}{6pt}
\renewcommand{\arraystretch}{1.08}
\begin{tabular}{@{}lcccc@{}}
\toprule
Method & WebShop & ALFWorld & ScienceWorld & Sokoban \\
\midrule
Base & $0.05\pm0.09$ & $26.35\pm1.78$ & $7.37\pm0.87$ & $7.35\pm1.13$ \\
GRPO-RLVR & $\mathbf{67.32\pm1.17}$ & $\mathbf{69.72\pm2.32}$ & $27.56\pm1.45$ & $58.33\pm0.55$ \\
CD (keep exp) & $65.86\pm0.98$ & $68.12\pm1.89$ & $27.20\pm1.03$ & $60.23\pm0.65$ \\
\rowcolor[gray]{0.95} CD (Ours) & $66.38\pm1.09$ & $67.58\pm2.05$ & $\mathbf{28.14\pm2.35}$ & $\mathbf{65.32\pm0.98}$ \\
\bottomrule
\end{tabular}
\end{table}

\paragraph{Pass@$128$.}
For each test instance and checkpoint, we generate a fixed pool of 256 independent rollouts.
We randomly select 128 rollouts per instance and compute Pass@$128$ as defined in Section~\ref{sec:metrics}, repeating the subsampling with 32 random seeds.
Table~\ref{tab:pass128_variability} reports the mean and sample SD across these 32 estimates, which reuse the same pools rather than generating new rollouts for every seed.

\begin{table}[!ht]
\centering
\caption{Pass@$128$ variability on Qwen2.5-7B-Instruct. We report mean $\pm$ SD (\%) over 32 subsamples from fixed pools of 256 rollouts per test instance.}
\label{tab:pass128_variability}
\vspace{1mm}
\small
\setlength{\tabcolsep}{6pt}
\renewcommand{\arraystretch}{1.08}
\begin{tabular}{@{}lcccc@{}}
\toprule
Method & WebShop & ALFWorld & ScienceWorld & Sokoban \\
\midrule
Base & $5.33\pm0.75$ & $78.33\pm0.56$ & $46.13\pm1.08$ & $45.12\pm0.79$ \\
GRPO-RLVR & $85.29\pm0.32$ & $91.22\pm0.35$ & $56.88\pm0.80$ & $68.94\pm0.81$ \\
\rowcolor[gray]{0.95} CD (Ours) & $\mathbf{88.98\pm0.22}$ & $\mathbf{94.10\pm0.31}$ & $\mathbf{62.75\pm0.88}$ & $\mathbf{78.83\pm0.37}$ \\
\bottomrule
\end{tabular}
\end{table}

\paragraph{Statistical scope and original results.}
For both tables, sample SD is computed across 32 scores with denominator $31$.
These SDs describe evaluation-time variation at fixed checkpoints, not variation across independent training runs; the Pass@$128$ SD is conditional on the generated rollout pools.
The original results in Tables~\ref{tab:main_results} and~\ref{tab:grpo_training_ceiling} remain point estimates and are not replaced by these supplementary means.
In the original Pass@$128$ evaluation, we sample up to 128 rollouts per instance and stop at the first success, which does not change its binary Pass@$128$ outcome.
The stored rollouts therefore do not form 128 complete Pass@$1$ evaluations; the 32 Pass@$1$ evaluations and the full 256-rollout pools above are collected separately.

\clearpage
\subsection{Example Prompts}
\begin{tcolorbox}[colback=gray!10, colframe=gray!40, title=EarlyExp Example Prompt]
\label{ee-prompt}
\textbf{User:}

You are an expert agent operating in the ALFRED Embodied Environment. Your task is to: put two cellphone in bed. Prior to this step, you have already taken 7 step(s). Below are the most recent 5 observations and the corresponding actions you took: 

[Observation 3: 'You arrive at desk 1...', Action 3: 'take cellphone 2 from desk 1']

[Observation 4: 'You pick up the cellphone 2 from the desk 1.', Action 4: 'go to bed 1']

[Observation 5: 'You arrive at bed 1. On the bed 1, you see ...', Action 5: 'go to desk 1']

[Observation 6: 'You arrive at desk 1. On the desk 1, you see ...', Action 6: 'take cellphone 1 from desk 1']

[Observation 7: 'Nothing happens.', Action 7: 'move cellphone 2 to bed 1']

You are now at step 8 and your current observation is: Nothing happens.
Your admissible actions of the current situation are: ['examine cellphone 2', 'examine desk 1', 'go to bed 1', 'go to drawer 1', 'go to drawer 2', 'go to garbagecan 1', 'go to shelf 1', 'go to shelf 2', 'go to shelf 3', 'go to shelf 4', 'go to shelf 5', 'go to shelf 6', 'go to shelf 7', 'go to shelf 8', 'go to sidetable 1', 'inventory', 'look', 'move cellphone 2 to desk 1'].

\par\medskip

The next action is **move cellphone 2 to desk 1**. Your task is to predict the environment observation after executing this action. Output only the observation.

\par\medskip

Assistant:

You move the cellphone 2 to the desk 1...
\end{tcolorbox}

\begin{tcolorbox}[colback=gray!10, colframe=gray!40, title=Counterfactual Distillation Example Prompt]
\label{Counterfactual Distillation-prompt}

\textbf{User:}

[STEP 1]
Observation: This room is called the bathroom. In it, you see...
Action: go kitchen

[STEP 2]
Observation: You move to the kitchen...
Action: pick up apple

[STEP 3]
Observation: You move the apple to the inventory...
Action: go to the counter

[STEP 4]
Observation: It's not clear how to go to/through a counter...
Action: go stove

[STEP 5]
Observation: It's not clear how to go to/through a stove...
Action: go counter

[STEP 6]
Observation: It's not clear how to go to/through a counter...
Action: focus on apple

\texttt{...}

You are an expert agent playing ScienceWorld, a text-based science simulation game.
Your task is to boil apple juice. For compounds without a boiling point, combusting the substance is also acceptable.

You have already taken 6 step(s), but the task failed without completing the objective.

FAILURE REASON: The environment closed early, but the task was not properly completed.

\par\medskip 

IMPORTANT:
Analyze your failed attempt and identify where things went wrong. You have a chance to go back to a specific step and try a different approach, with the benefit of your current experience.

First, carefully analyze the trajectory within \texttt{<think> </think>} tags. 

Then, provide your final decision in the following format:
\texttt{<step>} ... \texttt{</step>}
\texttt{<experience>} ... \texttt{</experience>}

\par\medskip

\textbf{Assistant:}

\texttt{<think>}
The trajectory initially seemed promising as I picked up the apple and moved towards the counter, but I became confused about how to proceed...
\texttt{</think>}

\texttt{<step>}2\texttt{</step>}

\texttt{<experience>} Plan the full procedure before acting: focus on the target substance, use valid appliances/containers, and avoid actions that the environment rejects (e.g., going ``to/through'' non-locations). \texttt{</experience>}

\end{tcolorbox}

\section{Examples}

\begin{figure*}
    \centering
    \includegraphics[width=0.7\linewidth]{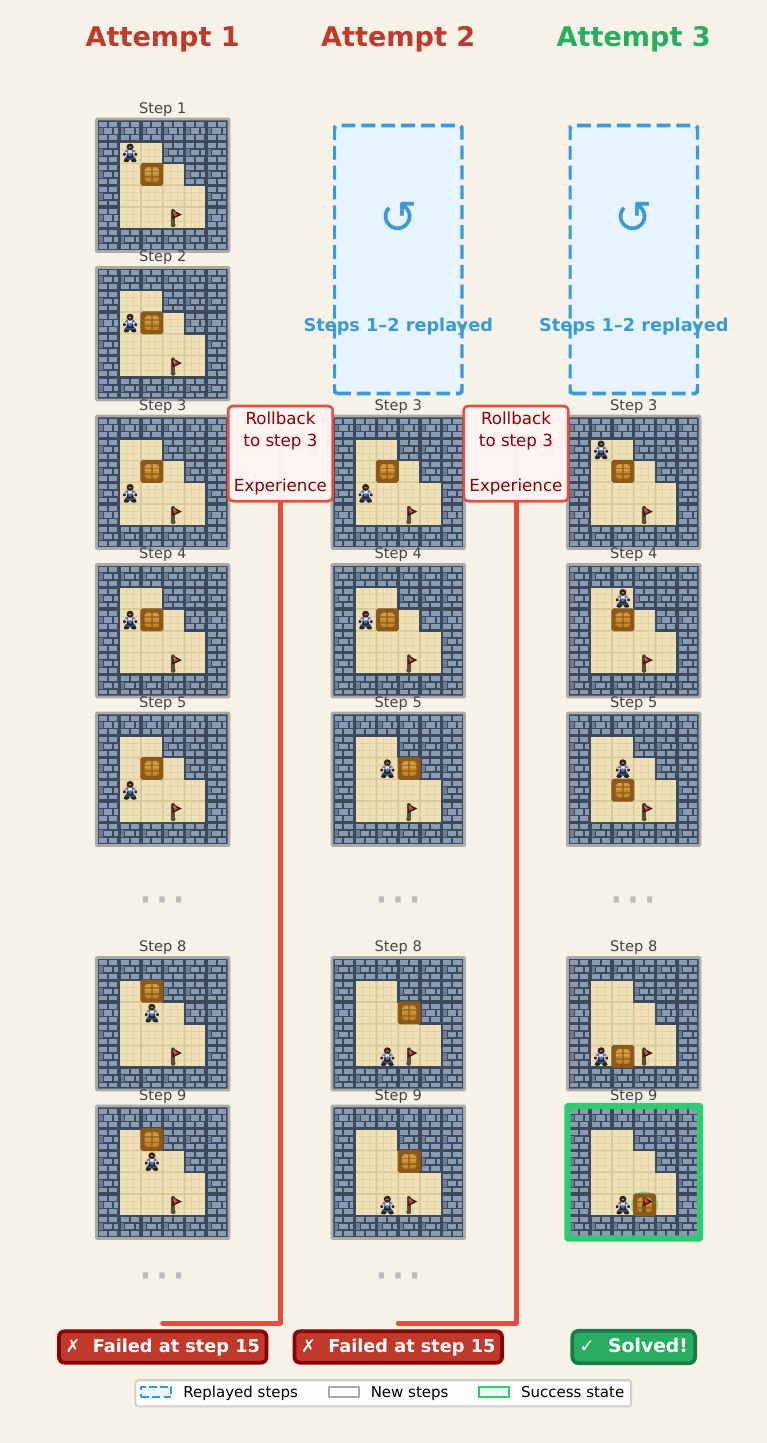}
    \caption{An example on Sokoban illustrating Stage 1 of Counterfactual Distillation. Starting from a failed trajectory, the agent reflects on the interaction history, identifies an earlier suboptimal decision(step 3), and generates a compact experience summary for rollback-based revision. The environment is then reset to the selected step, the prior history is replayed(step 1-2), and a new branch is explored under the guidance of the reflected experience. Repeating this failure → reflection → rollback → correction process enables the agent to recover from early mistakes and eventually reach a successful solution.}
    \label{fig:lefe-example}
\end{figure*}

\begin{figure*}
    \centering
    \includegraphics[width=0.7\linewidth]{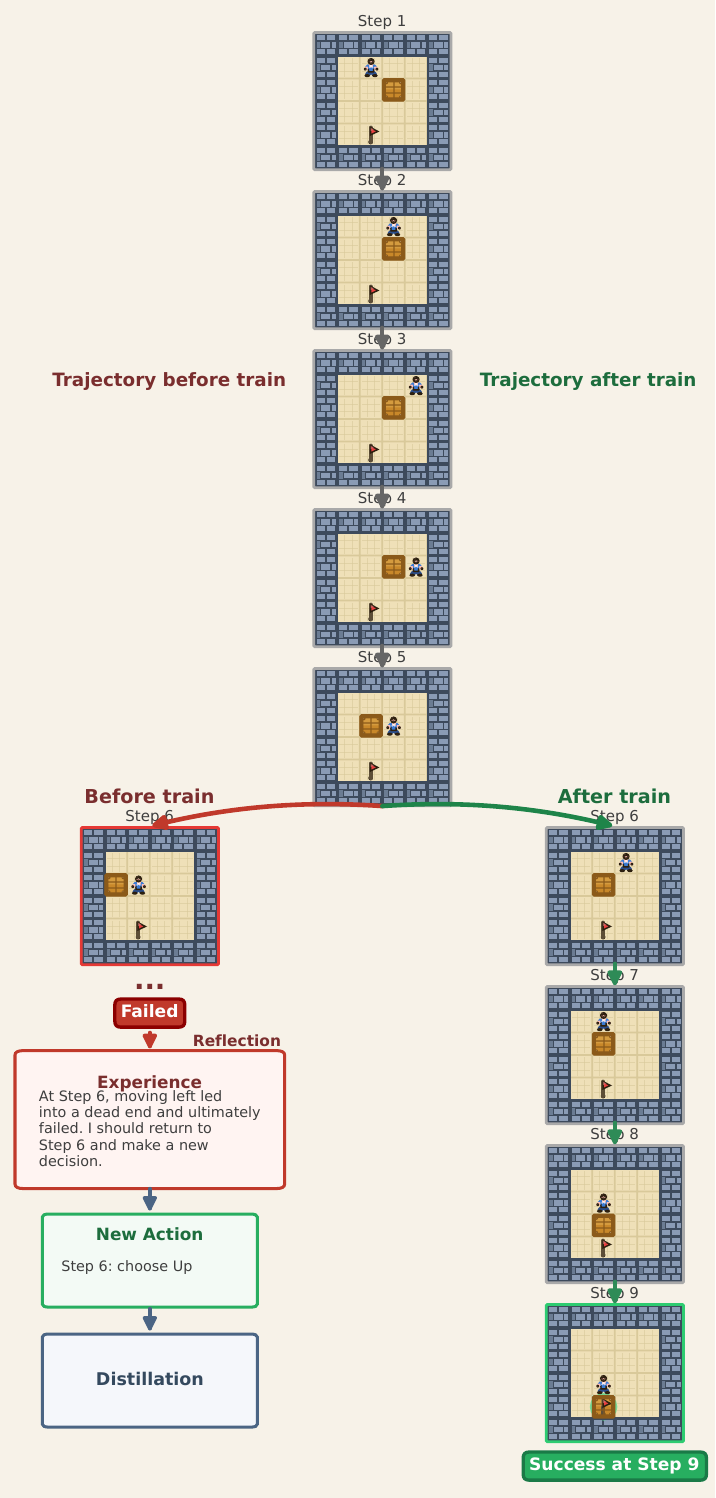}
    \caption{An example showing how Counterfactual Distillation distills experience-guided corrections into the model. Before training, the agent makes a suboptimal move at Step 6 and fails. Reflection identifies the error and suggests an alternative action, which is used as distillation supervision. After training, the model avoids the same mistake even without explicit experience at inference time, producing a successful trajectory at Step 9.}
    \label{fig:lefe-after-train-example}
\end{figure*}

\newpage
\section*{NeurIPS Paper Checklist}

\begin{enumerate}

\item {\bf Claims}
    \item[] Question: Do the main claims made in the abstract and introduction accurately reflect the paper's contributions and scope?
    \item[] Answer: \answerYes{} %
    \item[] Justification: The abstract and introduction clearly state the paper’s motivation, scope, and main contributions. These claims are supported by the proposed framework and the empirical results in Sections 3–4.

\item {\bf Limitations}
    \item[] Question: Does the paper discuss the limitations of the work performed by the authors?
    \item[] Answer: \answerYes{} %
    \item[] Justification: Yes. The paper includes a dedicated Limitations section that explicitly discusses the limitations of our method.

\item {\bf Theory assumptions and proofs}
    \item[] Question: For each theoretical result, does the paper provide the full set of assumptions and a complete (and correct) proof?
    \item[] Answer: \answerYes{} %
    \item[] Justification: Yes. Section 3 presents the formal setup, assumptions, notation, and training objectives of our method, providing the theoretical formulation of our proposal.

    \item {\bf Experimental result reproducibility}
    \item[] Question: Does the paper fully disclose all the information needed to reproduce the main experimental results of the paper to the extent that it affects the main claims and/or conclusions of the paper (regardless of whether the code and data are provided or not)?
    \item[] Answer: \answerYes{} %
    \item[] Justification: Sections~\ref{sec:cd}--\ref{sec:experiments} and Appendix~\ref{implementation-detals} describe our method and experimental protocols. Our implementation is available as open-source code.

\item {\bf Open access to data and code}
    \item[] Question: Does the paper provide open access to the data and code, with sufficient instructions to faithfully reproduce the main experimental results, as described in supplemental material?
    \item[] Answer: \answerYes{} %
    \item[] Justification: Our code is openly available, and Section~\ref{sec:experiments} and Appendix~\ref{implementation-detals} describe the datasets and experimental protocols.

\item {\bf Experimental setting/details}
    \item[] Question: Does the paper specify all the training and test details (e.g., data splits, hyperparameters, how they were chosen, type of optimizer) necessary to understand the results?
    \item[] Answer: \answerYes{} %
    \item[] Justification: Section~\ref{sec:experiments} and Appendix~\ref{implementation-detals} describe the training and evaluation settings; our implementation is also available as open-source code.

\item {\bf Experiment statistical significance}
    \item[] Question: Does the paper report error bars suitably and correctly defined or other appropriate information about the statistical significance of the experiments?
    \item[] Answer: \answerYes{} %
    \item[] Justification: Appendix~\ref{app:eval_variability} (Tables~\ref{tab:pass1_variability} and~\ref{tab:pass128_variability}) reports mean $\pm$ sample SD on four agent benchmarks with Qwen2.5-7B-Instruct, using 32 complete Pass@$1$ evaluations and 32 Pass@$128$ subsamples of fixed rollout pools. These SDs quantify evaluation-time variation at fixed checkpoints, rather than variation across training runs.

\item {\bf Experiments compute resources}
    \item[] Question: For each experiment, does the paper provide sufficient information on the computer resources (type of compute workers, memory, time of execution) needed to reproduce the experiments?
    \item[] Answer: \answerYes{} %
    \item[] Justification: Yes. We provide the necessary compute resource details in the appendix, including the type of compute workers, memory requirements, and execution time/cost for reproducing the experiments.
    
\item {\bf Code of ethics}
    \item[] Question: Does the research conducted in the paper conform, in every respect, with the NeurIPS Code of Ethics \url{https://neurips.cc/public/EthicsGuidelines}?
    \item[] Answer: \answerYes{} %
    \item[] Justification: Yes. We have reviewed the NeurIPS Code of Ethics, and the research conforms to it; our experiments use standard benchmark environments and do not involve human subjects or private data collection.

\item {\bf Broader impacts}
    \item[] Question: Does the paper discuss both potential positive societal impacts and negative societal impacts of the work performed?
    \item[] Answer: \answerNA{} %
    \item[] Justification: This work is a foundational method study evaluated on standard benchmark environments, and does not collect private data, or release high-risk data or models with direct societal impact.
    
\item {\bf Safeguards}
    \item[] Question: Does the paper describe safeguards that have been put in place for responsible release of data or models that have a high risk for misuse (e.g., pre-trained language models, image generators, or scraped datasets)?
    \item[] Answer: \answerNA{} %
    \item[] Justification: This work is a foundational method study evaluated on standard benchmark environments, and does not collect private data, or release high-risk data or models with direct societal impact.

\item {\bf Licenses for existing assets}
    \item[] Question: Are the creators or original owners of assets (e.g., code, data, models), used in the paper, properly credited and are the license and terms of use explicitly mentioned and properly respected?
    \item[] Answer: \answerYes{} %
    \item[] Justification: We cite the datasets, models, and codebases used in this work and respect their licenses and terms of use. Our implementation is available as open-source code.

\item {\bf New assets}
    \item[] Question: Are new assets introduced in the paper well documented and is the documentation provided alongside the assets?
    \item[] Answer: \answerYes{} %
    \item[] Justification: We release open-source code for our method; its data construction, training, and evaluation procedures are described in Sections~\ref{sec:cd}--\ref{sec:experiments} and Appendix~\ref{implementation-detals}.

\item {\bf Crowdsourcing and research with human subjects}
    \item[] Question: For crowdsourcing experiments and research with human subjects, does the paper include the full text of instructions given to participants and screenshots, if applicable, as well as details about compensation (if any)? 
    \item[] Answer: \answerNA{} %
    \item[] Justification: The paper does not involve crowdsourcing nor research with human subjects.

\item {\bf Institutional review board (IRB) approvals or equivalent for research with human subjects}
    \item[] Question: Does the paper describe potential risks incurred by study participants, whether such risks were disclosed to the subjects, and whether Institutional Review Board (IRB) approvals (or an equivalent approval/review based on the requirements of your country or institution) were obtained?
    \item[] Answer: \answerNA{} %
    \item[] Justification: The paper does not involve crowdsourcing nor research with human subjects.

\item {\bf Declaration of LLM usage}
    \item[] Question: Does the paper describe the usage of LLMs if it is an important, original, or non-standard component of the core methods in this research? Note that if the LLM is used only for writing, editing, or formatting purposes and does \emph{not} impact the core methodology, scientific rigor, or originality of the research, declaration is not required.
    \item[] Answer: \answerYes{} %
    \item[] Justification: LLMs are core experimental agents/policies; no private data; models listed in \S\ref{sec:experiments}

\end{enumerate}

\end{document}